\documentclass[Afour,sagev,times]{sagej}

\usepackage{moreverb,url}

\usepackage[colorlinks,bookmarksopen,bookmarksnumbered,citecolor=red,urlcolor=red]{hyperref}

\newcommand\BibTeX{{\rmfamily B\kern-.05em \textsc{i\kern-.025em b}\kern-.08em
T\kern-.1667em\lower.7ex\hbox{E}\kern-.125emX}}
\usepackage{siunitx}
\usepackage{amsmath}
\usepackage{amsfonts}
\usepackage{amssymb}
\usepackage{bm}
\usepackage{scalerel}

\usepackage{graphicx}
\usepackage{standalone}
\usepackage[table]{xcolor}
\usepackage{tikz}
\usetikzlibrary{shapes, backgrounds, shapes.geometric, arrows, arrows.meta, bending, positioning}
\usepackage{circuitikz}
\usepackage{pgfplots}
\usepackage{pgfplotstable}
\usepackage{ifthen} 
\usepackage{datatool}
\usepgfplotslibrary{groupplots, statistics, fillbetween}
\pgfplotsset{compat=newest}

\usepackage{subcaption}
\usepackage{multirow}%
\usepackage{amsthm}%
\usepackage{mathrsfs}%
\usepackage[title]{appendix}%
\usepackage{textcomp}%
\usepackage{manyfoot}%
\usepackage{booktabs}%
\usepackage{algorithm}%
\usepackage{algorithmicx}%
\usepackage{algpseudocode}%
\usepackage{listings}%
\usepackage{nicefrac}%

\pgfplotsset{
    /pgfplots/layers/Bowpark/.define layer set={
        axis background,
        axis grid,
        pre main, 
        main,
        post main, 
        axis ticks,
        axis lines,
        axis tick labels,
        axis descriptions,
        axis foreground
    }{/pgfplots/layers/standard},
}

\usepackage{media9}

\usetikzlibrary{external}
\def\app#1#2{%
  \mathrel{%
    \setbox0=\hbox{$#1\sim$}%
    \setbox2=\hbox{%
      \rlap{\hbox{$#1\propto$}}%
      \lower1.1\ht0\box0%
    }%
    \raise0.25\ht2\box2%
  }%
}

\begin{document}

\runninghead{Schumann \emph{et al.}}

\title{Resolving space-sharing conflicts in road user interactions through uncertainty reduction: An active inference-based computational model}

\author{Julian F. Schumann\affilnum{1}, Johan Engström\affilnum{2}, Ran Wei\affilnum{2}, Shu-Yuan Liu\affilnum{2}, Jens Kober\affilnum{1}, Arkady Zgonnikov\affilnum{1}}

\affiliation{\affilnum{1}Department of Cognitive Robotics, Delft University of Technology, Netherlands\\
\affilnum{2}Waymo LLC, Mountain View, CA, USA}

\corrauth{Johan Engström, Waymo LLC, 1600 Amphitheatre Parkway, Mountain View, California, 94043, USA}

\email{jengstrom@waymo.com}



\begin{abstract}
Understanding how road users resolve space-sharing conflicts is important both for traffic safety and the safe deployment of autonomous vehicles. While existing models have captured specific aspects of such interactions (e.g., explicit communication), a theoretically-grounded computational framework has been lacking. 
In this paper, we extend a previously developed active inference–based driver behavior model to simulate interactive behavior of two agents. Our model captures three complementary mechanisms for uncertainty reduction in interaction: (i) implicit communication via direct behavioral coupling, (ii) reliance on normative expectations (stop signs, priority rules, etc.), and (iii) explicit communication. In a simplified intersection scenario, we show that normative and explicit communication cues can increase the likelihood of a successful conflict resolution. However, this relies on agents acting as expected; in situations where another agent (intentionally or unintentionally) violates normative expectations or communicates misleading information, reliance on these cues may induce collisions. These findings illustrate how active inference can provide a novel framework for modeling road user interactions which is also applicable in other fields. 
\end{abstract}



\maketitle
\section{Introduction}

Understanding interactions between road users is an important aspect of road safety~\cite{ljung_aust_fatal_2012, habibovic2013driver} and key to the safe and efficient deployment of autonomous vehicles~\cite{domeyer2020vehicle,lee2026editorial}. In recent years, the topic has been a key focus of road user behavior research, which has resulted in a significant body of empirical findings as well as conceptual and computational models for understanding the key cognitive and behavioral mechanisms underlying road user interaction. 

Following the conceptual framework outlined by Markkula et al.~\cite{markkula_defining_2020}, we adopt their definition of a \emph{space-sharing conflict} as ``\emph{an observable situation from which it can be reasonably inferred that two or more road users are intending to occupy the same region of space at the same time in the near future}'' (page 736). Thus, a space-sharing conflict represents a situation where there is potential for road users to end up in a collision, given the uncertainty about how the situation will play out, but does not necessarily have to involve an imminent collision course, as in the traditional definition of a traffic conflict~\cite{amundsen1977proceedings}. Based on this, Markkula et al.~\cite{markkula_defining_2020} further define a (traffic) \emph{interaction} as a ``\emph{situation where the behaviour of at least two road users can be interpreted as being influenced by a space-sharing conflict between the road users}'' (page 737). 

Space-sharing conflicts can generally be resolved by reducing uncertainty~\cite{portouli2014drivers} and establishing a shared understanding, or common ground~\cite{clark1991grounding}, of the situation~\cite{markkula_defining_2020}. This can be achieved by various means of communication, where a general distinction can be made between implicit and explicit communication~\cite{sadigh_planning_2018}. Markkula et al.~\cite{markkula_defining_2020} define \emph{implicit communication} as ``\emph{... behavior which affects own movement or perception, but which can at the same time be interpreted as signaling something to or requesting something from another road user.}'' (page 741). Examples of implicit communication involve early yielding at an intersection to signal intent and looking towards another road user to indicate that they have been perceived. Means for implicit communication specifically related to movement and distance have been referred to as kinesics and proxemics respectively~\cite{domeyer2019proxemics,domeyer2020vehicle}. By contrast, \emph{explicit communication} can be defined as ``... \emph{behavior which does not affect own movement or perception, but which can be interpreted as signaling something to or requesting something from another road user.}'' (Markkula et al.~\cite{markkula_defining_2020}, page 742). Examples of explicit communication include using verbal utterances, hand gestures, vehicle lights, turn indicators, and the horn~\cite{de2019external}. Empirical results indicate that implicit signaling is the dominant means for communication while explicit communication happens more rarely~\cite{markkula_defining_2020,lee2021road}.

Another key feature of road user interaction is interdependence~\cite{domeyer2020interdependence,noonan2022interdependence} or coupling~\cite{domeyer2022driver} where the behavior of one actor depends reciprocally on the behavior of another actor. However, as shown by Domeyer et al.~\cite{domeyer2019proxemics} and Noonan et al.~\cite{noonan2022interdependence}, such interdependent behavior appears to mainly occur in uncontrolled scenarios, whereas the behaviors of road users tend to be less dependent in scenarios where interactions are facilitated by traffic signals or signs.
This suggests that traffic controls, such as traffic lights and stop signs, and the associated road rules reduce the need for direct, interdependent communication between road users. 

A wide range of computational models have been proposed to explain these phenomena. This includes models based on game theory~\cite{camara2018towards,camara_pedestrian_2021,fox2018should,fisac2019hierarchical,kang2017game, schwarting_social_2019, thalya2020modeling, sadigh_planning_2018}, evidence accumulation~\cite{boda2020computational, giles2019zebra, markkula_evidence_2018, pekkanen_variable-drift_2022, zgonnikov_should_2024, markkula_explaining_2023}, coupled dynamical systems~\cite{domeyer2020interdependence,noonan2022interdependence}, social force models~\cite{johora2020zone}, and the communication-enabled interaction model~\cite{siebinga_model_2024-1}. 
These models bring different perspectives and have significantly advanced our understanding of road user interactions. Some models, such as the the social force model~\cite{johora2020zone} or Siebinga \emph{et al.}'s communication-enabled interaction model~\cite{siebinga_model_2024-1,siebinga2023modelling}, are formulated as general models, applicable across multiple scenarios such as pedestrian crossings or highway merging. Nevertheless, such existing models do not offer an integration into a more general theory of human cognition and behavior. Importantly, none of these models implemented either explicit communication or traffic norms.

In this paper, we explore \emph{active inference} as a framework for modeling the collective behavior of human road users in resolving space-sharing conflicts in traffic interactions. In the spirit of recent work on active inference modeling of interactive behavior in other domains, we assume that road users maintain reciprocal beliefs about each other’s future behavior as the basis for selecting their next actions (e.g., braking to yield or accelerating to go first when crossing an intersection). As long as the interacting agents have a shared \emph{and precise} understanding about how the situation will play out (e.g., who will go first), the space-sharing conflict will be resolved smoothly, in a similar way as when two agents are singing a familiar song together~\cite{friston2015active,friston2015duet} or performing other joint actions such as dancing or jointly carrying an object~\cite{pezzulo_predictive_2026}. In such situations, the agents’ beliefs and ensuing actions can be seen as guided by a common, agent-neutral, \emph{generative model}~\cite{pezzulo_predictive_2026}, typically established through implicit communication (e.g., via proxemics and kinesics~\cite{domeyer2019proxemics,domeyer2020vehicle}). 

However, in some traffic situations a sufficiently precise common understanding may be lacking due to uncertainty in the agents’ beliefs (e.g., agent $A$ is uncertain about whether agent $B$ will yield), which may preclude efficient resolution of the space-sharing conflict. We here propose two main ways in which uncertainty in road user interaction may be resolved -- relying on established \emph{normative expectations}~\cite{bicchieri2016norms, fraade2025being} and \emph{explicit communication} -- and demonstrate how these two mechanisms can be represented in a computational mechanistic road user behavior model.

In this framing, normative expectations can be seen as a form of \emph{niche construction}~\cite{constant2018variational,bruineberg2018free} where traffic rules and more informal social norms have been established in a traffic culture to guide, or scaffold, expectations about road user behavior, supported by deontic cues~\cite{constant2019regimes} such as road signs, traffic signals and lane markings. Explicit communication can be understood as an \emph{epistemic action}~\cite{friston_active_2017, parr_active_2022, pezzulo_predictive_2026} with the purpose to obtain new information that could potentially reduce uncertainty in the beliefs about the other agent’s intent. 

We implement these mechanisms into an active inference model previously proposed for capturing the behavior of individual drivers in non-interactive scenarios~\cite{engstrom_resolving_2024,schumann_active_2025}. We then use this extended model to show how uncertainty in human interactions can be reduced through (i) implicit communication -- a baseline approach (modeled similarly to Siebinga \emph{et al.}~\cite{siebinga_model_2024-1}), (ii) explicit communication~\cite{vasil_world_2020} and (iii) normative expectations~\cite{bicchieri2016norms, fraade2025being} and explicit deontic cues such as traffic lights, painted road markings and stop signs~\cite{constant2019regimes,laurent_traffic_2021}. We evaluate the model in a simplified intersection scenario (Figure~\ref{fig:overview}) with two interacting vehicles, similar to Fox \emph{et al.}~\cite{fox2018should}. We demonstrate how norms and communication can (jointly or separately) facilitate successful coordination, illustrating  how active inference can unify behavioral coupling (through implicit communication), explicit communication, and normative reasoning within a single modeling framework. Furthermore, we explore how miscommunication or failures to adhere to normative expectations can lead to conflicts and collisions under those principles.

\section{Method}
\subsection{Model overview}
Our model is grounded in active inference -- a general framework for understanding and modeling sentient behavior of living systems~\cite{friston_active_2017,parr_active_2022}. It is based on the key idea that agents strive to sustain their existence by forming (probabilistic) beliefs about how observations are caused by the environment and their own actions, and pursuing policies (action sequences) that generate observations that are preferred or expected given the type of creature they are (maximizing \emph{pragmatic value}) or allow the agent a better understanding of their environment (maximizing \emph{epistemic value}). Observations that are not aligned with the preferred expectations are surprising to the agent and, in order to be in tune with its environment, the agent strives to minimize surprise over time, which can be modeled in terms of the minimization of variational free energy. When acting in an uncertain environment, such as road traffic, a key aspect in minimizing surprise or free energy is to control the uncertainty in one’s beliefs about how the situation will play out, for example, whether another road user will yield or cross at an intersection ahead.

We base our model on a framework previously developed for active-inference modeling of human driver behavior~\cite{engstrom_resolving_2024, schumann_active_2025}. Fundamental to active inference, this framework includes a global environment model (i.e., the \emph{generative process}) as well as each agent's internal \emph{generative model} approximating this environment. The environment is described by the state $\bm{\eta}$, from which agents make observations $\bm{o}$ according to the \emph{generative process}'s observation probability $\widehat{p}(\bm{o}\mid \bm{\eta})$. The state of the environment can be influenced by the agents' combined actions $\bm{a}$ according to the \emph{generative process}'s state transition function $\widehat{p}(\bm{\eta}'\mid \bm{\eta}, \bm{a})$. Meanwhile, each agent's \emph{generative model} is based on an internal state $\bm{s}$, as well as the corresponding observation probability $p(\bm{o}\mid \bm{s})$ and a state transition function $p(\bm{s}'\mid \bm{s}, \bm{a}_v)$. A central tenet of active inference is that the agent does not know the exact state of the world but rather maintains an internal probabilistic belief $q(\bm{s})$ over $\bm{s}$). While it is possible to assign different generative models to different agents, in this work, both agents use separate instances of the same model (i.e., their understanding about the world as well as their preferences priors for desired observations are identical).

\begin{figure}
    \centering
    \includegraphics[width = \linewidth]{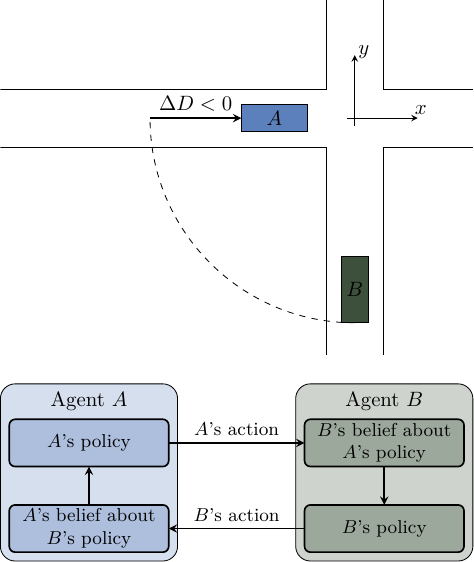}
    \caption{The modeled scenario: two agents on crossing paths representing a space-sharing conflict with $\Delta D$ denoting agent $B$’s lead over agent $A$ (i.e., $\Delta D < 0$ if A is closer to the intersection). Below it, the basic action-perception (hermeneutic) cycle resulting from modeling two active inference agents observing each other (based on Friston and Frith~\cite{friston2015active})}
    \label{fig:overview}
\end{figure}

In our model, each of the modeled agents from a set of agents $V$ repeatedly chooses an action based on a multi-step process, following the model of Schumann \emph{et al.}~\cite{schumann_active_2025} (see Appendix for a more detailed mathematical description):
\begin{enumerate}
    \item The agent $v\in V$ makes an observation $\bm{o}$ and uses it to update its belief $q(\bm{s})$, which includes both its own state and that of the other agents.
    \item The agent generates beliefs about the possible future kinematic states of the other agent, representing the uncertainty about where the other agent may move next.
    \item The agent generates candidate policies $\bm{\pi}$, that is, sequences of future actions $\bm{a}_v$ up to a planning horizon. In the exiting model~\cite{schumann_active_2025}, the actions comprised acceleration and steering rate (in the current model we add explicit communication signals as another type of action, as further described below). Using the \emph{generative model's} state transition function, the agent's beliefs about the future states associated with each policy $\widehat{q}(\bm{s})$ are generated.
    \item Candidate policies are evaluated based on the expected free energy (EFE) associated with the beliefs about the future consequences of adopting a given policy, and the policy with the lowest EFE is selected. The EFE scores both the value of achieving an agent's goals as implemented in the preference prior $p(\bm{o})$ (maximizing \emph{pragmatic value}) and the value of obtaining new information to resolve uncertainty (maximizing \emph{epistemic value}).
    \item The agent only updates its currently pursued policy if the accumulation of surprising evidence $E$ against the current policy has reached a predefined threshold. Otherwise, it extends the current \emph{reference} policy by appending a single action at the end of the time horizon. This mechanism allows the model to capture realistic reaction times of human drivers~\cite{schumann_active_2025}. In practice, the accumulated surprise is linearly dependent on the \emph{pragmatic value} (i.e., a low pragmatic value drives the agent's need for a comprehensive policy update, which is also referred to as ``re-planning'' by Schumann \emph{et al.}~\cite{schumann_active_2025}). The first actions of the currently selected policies for each agent are then executed simultaneously, based on which the environment is updated, and new observations generated.
\end{enumerate}

The model of Schumann et al.~\cite{schumann_active_2025} used the above five steps to model human collision avoidance behavior in non-interactive safety-critical scenarios, where a modeled agent responded to another agent executing a predefined policy. Here, we extended this model to represent the dyadic interaction between two agents ($V = \{A,B\}$) on a crossing path at an intersection (Figure~\ref{fig:overview}, top panel) aimed at resolving the resulting space-sharing conflict. 

In such a scenario, our model implementation leads to the coupled action-perception hermeneutic cycle (Figure~\ref{fig:overview}, bottom panel) where each agent maintains, and repeatedly updates, its belief about the other agent's policy. To resolve the space-sharing conflict, the agents can adapt their behavior to one another in a reciprocal fashion, what we here refer to as \emph{direct coupling} (e.g., one agent slows down to yield while the other is accelerating to cross first in response), reducing the uncertainty in their beliefs about future collisions via \emph{implicit communication}. 

In the traffic interaction literature, implicit communication is often understood as \textit{intentional} signaling of intent through kinematics~\cite{tian2023deceleration,dey2017pedestrian,miller2022implicit,zgonnikov2024nudging}, for example, a vehicle slowing down early when approaching a zebra crossing to let the pedestrians know they can cross. However, here we use this term in a more general sense, implying communication that de-facto happens through vehicle kinematics regardless whether the agents intend changes (or lack of changes) in own kinematics as a communicative act. For instance, if a driver intends to merge onto a highway lane, a driver of a vehicle already in that lane would likely interpret initial movement of the merging vehicle towards the target lane as an indication of the forthcoming merge. Here, the kinematics of the merging vehicle effectively signals the intent of that vehicle's driver to surrounding road users even if that driver is not aware of their presence. This notion of implicit communication is consistent with the definition of Markkula et al.~\cite{markkula_defining_2020}: ``\emph{...behaviour which affects own movement or perception, but which can at the same time be interpreted as signalling something to or requesting something from another road user}'' (page 741).

Besides this direct coupling via implicit communication, where inferences about the other agents' intents are made purely based on their kinematics, other methods of uncertainty reduction are also possible. The first mechanism considered in this work is the reliance on \emph{normative expectations}~\cite{bicchieri2016norms, fraade2025being} -- shared beliefs within a society about how one should behave. Normative expectations may be encoded in formal rules such as the traffic code but may may also include more informal conventions. The second mechanism for reducing uncertainty about other agents' policies is \emph{explicit communication}, which enables an agent to directly signal its intent and thereby allows others to more accurately predict its future behavior. Both mechanisms can reduce uncertainty and therefore can potentially lead to a more reliable resolution of space-sharing conflicts. The following two subsections describe how normative expectations and explicit communication are implemented in our model.

\subsection{Modeling normative expectations}\label{sec:norm_condition}
In driving, an agent can rely on established conventions for interactions -- closely related to the notion of ``interaction rituals'' proposed by Goffman~\cite{goffman2017interaction} and discussed by Markkula \emph{et al.}~\cite{markkula_defining_2020} -- as a means for resolving space-sharing conflicts with other road users. Such conventions can be understood in terms of \emph{normative expectations}~\cite{bicchieri2016norms, fraade2025being}, that is, shared societal beliefs about proper road behavior. These expectations allow an agent to constrain its beliefs about the expected actions of other road users, thus reducing uncertainty and improving the likelihood of a successful interaction. In traffic, \emph{normative expectations} are partially encoded in formal road rules and partially embodied in informal conventions, and are often reinforced by \emph{deontic cues} such as designed road features such as lane markings, traffic lights, and stop signs~\cite{constant2019regimes}.

To implement these normative expectations into our active inference framework, we employ the \emph{norm-conditioned particle filter} introduced by Schumann \emph{et al.}~\cite{schumann_active_2025}. In this approach, the \emph{generative model's} state transition probability is no longer solely determined by kinematic likelihoods~\cite{engstrom_resolving_2024}, but is instead biased by the \emph{projected normative probability}. The projected normative probability captures the assumption that a modeled agent $v$ assigns low probability to actions by other road users that would -- if pursued consistently over short- and medium-term horizons (up to $\SI{4}{s}$ in our implementation) -- lead to norm violations (see Equation~\eqref{eq:state_norm} in the Appendix for mathematical details). Norms could then be specified via the \emph{normative probability} $p_n$, which for instance is low for a vehicle exiting its current lane unless the driver signaled their intent for doing so; in that case, when making predictions of other agents, the model of Schumann et al.~\cite{schumann_active_2025} would consequently assign a low \emph{projected normative probability} to large steering maneuvers by other vehicles if these would result in the other vehicle exiting their lane. 

Importantly, however, the norm-conditioned particle filter as implemented by Schumann \emph{et al.}~\cite{schumann_active_2025} can only account for ``non-interactive''  norm violations by a surrounding agent $w$. Such non-interactive violations depend solely on the violating agent's own behavior but not on other agents. Concretely, Schumann et al.~\cite{schumann_active_2025} assumed that
\begin{equation}
    p_n(\bm{s}) = \prod\limits_{w \in V\setminus \{v\}} p_{n,V}(\bm{s}_{w}) \,, \label{eq:normative_weight_old}
\end{equation}
where $\bm{s}_w$ denotes the state variables associated with agent $w$, such as position or velocity. Such non-interactive norms are represented, for instance, by lane following, adhering to speed limits, traffic lights, or stop signs. If not only the norms, but also the base kinematic likelihoods in the state transition probability are independent, the state transition function can be factorized between agents. This is computationally advantageous, as the future trajectories of all other agents need to be predicted only once when agent $v$ evaluates multiple policies. This is particularly beneficial in crowded scenes.

In our proposed model, however, we extend this formulation by adopting the more general assumption of \emph{interactive norms}, which means that the norm compliance of another road user's action can only be assessed in relation to the behavior of the ego (the modeled) agent. Consequently, we modified Equation~\eqref{eq:normative_weight_old}, yielding
\begin{equation}
    p_n(\bm{s}) = \prod\limits_{w \in V\setminus \{v\}} p_{n,V}(\bm{s}_{w} \vert \bm{s}_{v}) \,, \label{eq:normative_weight}
\end{equation}
with the \emph{normative probability} of each surrounding agent $w$ explicitly depending on the state of the modeled agent $v$. With such interactive norms our model can capture priority rules (which we investigated here), but potentially also other cases, such as rules mandating minimum following distances.

Crucially, in scenarios involving at least one interactive norm, the modeled agent $v$ can no longer be factorized out of the \emph{generative model's} state transition function. As a result, the future trajectories of other road users must be regenerated for each candidate policy, leading to a substantial increase in computational cost.

\subsubsection{Types of norms}
In this work, we implemented several types of norms. First, we re-used the lane-following norms of the existing model~\cite{schumann_active_2025}: the model expects other agents to stay in their lane and on the road. Second, in our model agents always assume that other traffic participants will follow a speed limit, avoiding velocities above $v = \SI{11}{m.s^{-1}}$ ($\SI{1}{m.s^{-1}}$ above the assumed posted speed limit in the current simulation, see below) if possible.

Third, we investigate the impact of additional traffic rules representing norms specifically related to interactions with other road users, henceforth referred to as \emph{interaction-specific norms}: the ``stopping at a stop sign'' rule and the ``first come, first go'' priority rule. The former is implemented as a requirement for an agent to fall below a velocity limit of $v_L = \SI{0.278}{m.s^{-1}}$ (i.e., $\SI{1}{km.h^{-1}}$) in a short distance in front of the stop sign. Concretely, states where another agent $w$ is moving beyond the stop sign without having previously stopped will result in lower values of that agent's normative weight function $p_{n,V}$ (see Equation~\eqref{eq:normative_weight}). 
For the priority rule, each agent $v$ considers the other agent $w$ to have priority if they are closer to the intersection. While this is initially reevaluated at every timestep, once agent $v$ completed the stop in front of the stop sign, its opinion about the priority becomes fixed. If an agent believes it has priority, the normative weight function $p_{n,V}$ (Equation~\eqref{eq:normative_weight}) will assign low likelihoods to observations where the other agent $w$ enters the intersection without trailing agent $v$ by a sufficient amount. 

In our model, each agent not only assumes that the other agent follows the above norms, but also tries to follow them itself. This is implemented via the model's prior preferences $p(o)$ (Equation~\eqref{eq:preference_likelihood}): policies leading to norm violations are disincentivized via low pragmatic values. 
The exact implementation of the norms and preference prior adjustments can be found in the Appendix.

\subsection{Modeling explicit communication} \label{sec:Communication}
Another way for agents to reduce the uncertainty about each other's policies is by using explicit communication. To implement this mechanism, we endow each agent with the capacity for two communicative acts $\bm{\gamma} = \{\gamma_{A}, \gamma_{Y}\}$:
\begin{itemize}
    \item $\gamma_A$: \textbf{Asking} (prompting) the other agent to communicate their intent, with the expectation of a reply.
    \item $\gamma_Y$: Signaling one's intention to \textbf{yield}, informing the other agent about one's intended behavior
\end{itemize}
For simplicity, we assume that the intended signal can be directly mapped from the communicative act (see~\cite{vasil_world_2020} for a more elaborate account of how such mappings are established in the first place). We also model these communicative acts at a high level of abstraction and do not attempt to represent their specific communication modalities. For example, in a real interaction, prompting may be mediated by seeking eye contact and yielding could be signalized with a hand gesture, but in the current model we abstract away from the detailed ways in which prompting and signaling yielding are instantiated.

These signal states are defined as discrete, binary variables (i.e., $\bm{\gamma} \in \{0,1\}^2$) -- with agents either signaling ($\gamma_A, \gamma_Y = 1$) or not ($\gamma_A, \gamma_Y = 0$) -- when they are part of the \emph{generative process}'s state $\bm{\eta}$, of the observations, or of the policies. However, in an agent's belief state $\bm{s}$, they are defined over a continuous domain (i.e., $\bm{\gamma}_s \in [0,1]^2$). 
This implementation of belief states as continuous variables has two purposes: (1) It captures perceptual uncertainty (e.g., due to distance between agents), allowing agents to represent uncertainty about whether a signal was observed. (2) Under the assumption of honest communication by the other agent $w$, the belief $q(\gamma_{Y,s,w})$ can be interpreted as agent $v$'s belief over agent $w$'s \emph{yielding intent}. Importantly, this belief $q(\gamma_{Y,s,w})$ is only updated based on the observed yielding signal $\gamma_{Y,o,w}$ in our current implementation. In reality though, agent $v$'s belief about agent $w$'s yielding intent would also be influenced by other sources of information (e.g., priority rules or kinematics). Consequently, here $\gamma_{Y,s,w}$ only represents the \emph{signalized} yielding intent of agent $w$, but for simplicity in the rest of the paper we refer to it simply as \emph{yielding intent}.

Furthermore, to allow agents to react to perceived signals, we select the signaling policy (i.e., prompting and yielding) at every timestep, in contrast to the evidence accumulation-based re-planning used for the kinematic policy (step 5 in the description of the existing model above). This allows for reaction to perceived signals even when the currently selected kinematic policy is considered sufficient. Additionally, to allow for more immediate actions, once a yielding signal is perceived, an agent updates its kinematic policy accordingly.

\subsubsection{Prompting for intent}
In our model, the prompting by agent $v$ for the yielding intent of another agent $w \in V \setminus \{v\}$ -- that is, to choose the action $\gamma_{A,a,v} = 1$ -- is driven by the \emph{epistemic value} component of the expected free energy. If agent $v$ does not prompt ($\gamma_{A,a,v} = 0$), the epistemic value is always approximately zero, reflecting the fact that no new information is expected to be gained.  
In contrast, when prompting (under the assumption that agent $w$ is cooperative), agent $v$'s belief over $w$'s yielding intent $q(\gamma_{Y,s,w})$, with $\gamma_{Y,s,w} \in [0,1]$, collapses to a quasi-binary state through the \emph{generative model}'s state transition function (Equation~\eqref{eq:signal_state_transition}). In particular, the observation model $p(\bm{o}\mid\bm{s})$ for $\gamma_{Y,w}$ follows a Bernoulli distribution $\mathcal{B}_{\gamma_{Y,s,w}}(\gamma_{Y,w})$, resulting in the belief $q(\gamma_{Y,s,w})$ concentrating around the extreme values $\gamma_{Y,s,w} = 0$ and $\gamma_{Y,s,w} = 1$.  
Under this belief, agent $v$ assumes that -- given a hypothetical future observation -- it will become certain about agent $w$'s yielding intent. This corresponds to the \emph{expected ambiguity} component of the epistemic value, representing the reliability of observations~\cite{friston_active_2017,parr_active_2022}, being zero.  (Equation~\eqref{eq:pragmatic_epistemic}). Consequently, the epistemic value of the prompting action ($\gamma_{A,a,v} = 1$) reduces to the \emph{posterior predictive entropy}~\cite{friston_active_2017,parr_active_2022} (Equation~\eqref{eq:pragmatic_epistemic}), which is maximized when the uncertainty about agent $w$'s yielding intent is highest (and hence a prompting action has the maximum potential to reveal new information about $w$' intent). Specifically, the epistemic value of the prompting action is highest when there is an initial belief with $\mathbb{E}_{q(\gamma_{Y,s,w})}\gamma_{Y,s,w} = 0.5$ (see Equation~\eqref{eq:g_prompt_full} for the corresponding derivation).

In addition, in the preference prior underlying the pragmatic value (Equation~\eqref{eq:preference_likelihood}), we implement a base cost for active signaling, as in real life this will involve some amount of effort. Therefore, an agent will only send a prompting signal if the epistemic value of that action outweighs the pragmatic cost associated with the effort of doing so.

\subsubsection{Signaling intent}
The agent's motivation for sending a yielding signal is grounded in the assumption that agents are cooperative. Although signaling yielding, as with prompting before, incurs a base cost representing the required effort, the penalty in the preference prior for uncooperative behavior is higher. Specifically, if an agent $v$ believes that another agent $w$ is prompting for its intent (i.e., $\gamma_{A,s,w} \rightarrow 1$) and simultaneously intends to yield (i.e., under its current policy, it predicts arriving second at the intersection), then failing to signal yielding (i.e., setting $\gamma_{Y,v}=0$) leads -- via the aforementioned added cost to the preference prior -- to a substantial decrease in \emph{pragmatic value}.

However, this pragmatic value penalty is not the primary effect of the yielding signal in our model. Instead, the main impact of the signal lies in reducing the uncertainty in the prompting agent $w$'s beliefs about the yielding agent $v$'s future behavior. This follows from the assumption introduced above that an agent's belief in having perceived a yielding signal directly translates into the belief that the other agent intends to yield.
From the perspective of agent $w$, therefore, perceiving a yielding signal from agent $v$ (\emph{i.e.}, we have $q(\gamma_{Y,s,v}) \approx \delta(\gamma_{Y,s,v}-1)$ in agent $w$'s belief) leads agent $w$ to expect that agent $v$ intends to yield. This expectation is implemented in the \emph{normative probability}, where behaviors of agent $v$ that are consistent with yielding are assigned higher probability. Concretely, under the expectation that agent $v$ intends to yield, predictions in which agent $v$ crosses first at the intersection are assigned low probability, biasing agent $w$ toward believing that agent $v$ will yield by decelerating. Consequently, as long as those required decelerations are reasonable, agent $w$ is confident that agent $v$ remains behind once agent $w$ reaches the intersection, implying that crossing the intersection is safe. In this way, a yielding signal can promote the successful resolution of the modeled interaction at the intersection. 
Additionally, if agent $v$ is perceived to have signaled yielding, agent $w$ will consider it acceptable to ignore priority rules (\emph{i.e.}, there will be now pragmatic cost for priority violations), with explicit communication in this case superseding normative expectations.

\subsection{Simulated scenario}
We simulated the model behavior in an interaction between two agents ($V = \{A,B\}$) on a crossing path at an intersection (Figure~\ref{fig:overview}, top panel). In this scenario, the two vehicles start at $t=0$ at respective distances of $D_A(0) = x_A(0) < 0$ and $D_B(0) = y_B(0) < 0$ -- defined based on their center of mass position. Specifically, we set that $D_A(0) = - \SI{65}{m}$, while varying $D_B(0) \in \{- \SI{90}{m},- \SI{80}{m},- \SI{70}{m},- \SI{68}{m},-\SI{66.5}{m},- \SI{65}{m}\}$, resulting in initial distance differences $\Delta D(0) = D_B(0) - D_A(0) \in \{- \SI{25}{m},- \SI{15}{m},- \SI{5}{m},- \SI{3}{m}, - \SI{1.5}{m},\SI{0}{m}\}$. Both vehicles start with an initial velocity of $v_1(0) = v_2(0) = \SI{10}{m.s^{-1}}$ towards the intersection. To capture the variability due to the stochasticity of our model, we simulated its behavior 50 times for each $\Delta D(0)$ value.

\section{Results}
In this section we first demonstrate how outcomes of the modeled interaction are affected by four different regimes: (1) the baseline using \emph{implicit communication}, (2) reliance on interaction-specific \emph{normative expectations}, (3) using \emph{explicit communication}, and (4) a combination of the latter two. We then zoom in on example simulations representing each regime in respective subsections. Finally, we demonstrate the effects of  miscommunication and overtrust in communication or norm-following.

\subsection{Interaction outcomes}
\begin{figure}
    \centering
    \includegraphics[]{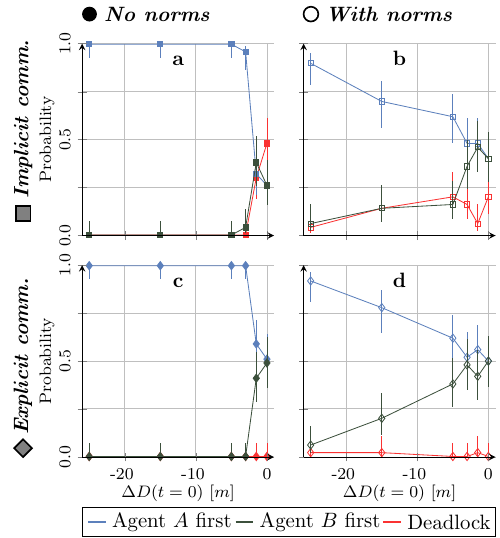}
    \caption{The likelihoods of different simulation outcomes under different model setups, for different initial differences in distance to the intersection $\Delta D(0)$. ``With norms'' refers to models considering both first-come priority rules and stop signs. The vertical lines indicate the 95\% confidence intervals of the shown probabilities according to the Wilson score~\cite{wilson1927probable}.}
    \label{fig:prob_results}
\end{figure}

Under the baseline regime (implicit communication only), whether a space-sharing conflict is resolved depends on the presence of asymmetries in agents' proxemics (Figure~\ref{fig:prob_results}\textbf{a}). Sufficiently large initial asymmetries (i.e., Agent $A$ having a significant lead of $\Delta D \leq \SI{-5}{m}$, which is equivalent to an initial time gap between agents of $\SI{0.5}{s}$) allow the leading agent to always go first. As the difference in initial distances decreases, however, other outcomes become more likely; for instance, for $\Delta D > \SI{-5}{m}$ the trailing agent goes first in some cases. For such small differences the conflict may not even be resolved at all, resulting in a deadlock in which both agents come to a standstill and fail to proceed through the intersection. For instance, for $\Delta D=0$, there is approximately a 489\% chance of such a deadlock.

As theorized above, our simulations in the controlled intersection scenario with norms revealed that normative expectations can reduce the likelihood of a deadlock for initially symmetric scenarios (Figure~\ref{fig:prob_results}\textbf{b}). Specifically, for $\Delta D > \SI{-5}{m}$, the likelihood of a deadlock falls to 20\%, compared to nearly every second case in the baseline scenario. Meanwhile, however, for cases where Agent $A$ had a larger lead, both deadlocks and role reversal (i.e., Agent $B$ going first) become possible.  

In the absence of normative expectations (i.e., uncontrolled intersection), explicit communication allowed the agents to successfully resolve every interaction (Figure~\ref{fig:prob_results}\textbf{c}). Notably, at small $\Delta D$ values the leading agent can yield priority to the lagging agent -- this is due to random variations in policy sampling; although this is not consistent with the initial conditions, this outcome is not implausible and avoids the deadlock. Thus,  we consider it a successful resolution of the conflict.

Although in the uncontrolled intersection case the addition of communication allows the agents to successfully resolve any interaction (Figure~\ref{fig:prob_results}\textbf{a} vs. Figure~\ref{fig:prob_results}\textbf{c}), controlled intersections with interaction-specific norms are very common in the real world, making them an interesting case for analyzing the effect of communication. 
Adding communication to the model equipped with normative expectations significantly reduces the likelihood of deadlocks across all starting conditions (Figure~\ref{fig:prob_results}\textbf{b} vs. Figure~\ref{fig:prob_results}\textbf{d}). For instance, for $\Delta D = 0$ this likelihood decreased from 20\% to 0\%. Nevertheless, by contrast to the scenario with only communication, a small number of deadlocks are still present. 

\subsection{Interaction dynamics}
To highlight the mechanisms at play, we analyzed the dynamics of representative simulations of our model in each of the four regimes.

\subsubsection{Baseline: Implicit communication}\label{sec:direct_coupling}
\begin{figure*}
    \centering
    \includegraphics[]{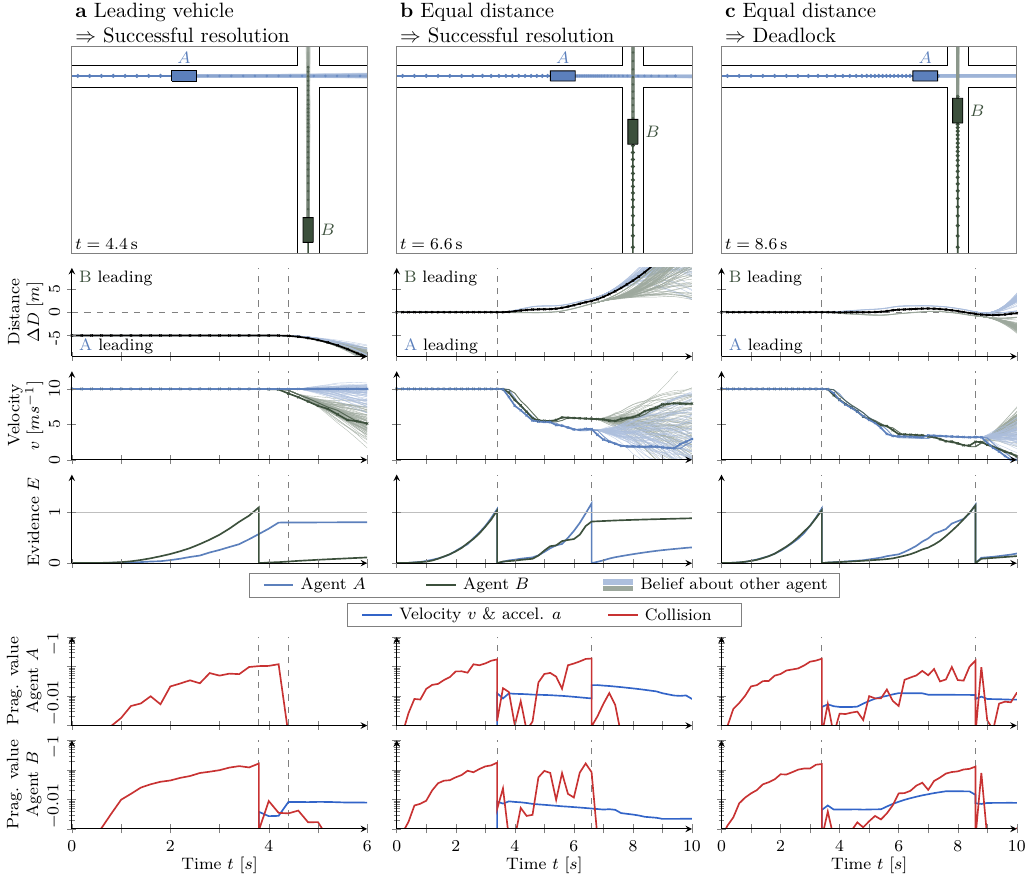}
    \caption{Three illustrative examples of model behavior in the baseline scenario: the agents interact based on implicit communication only. For each example simulation, the panels illustrate (from top to bottom) the top-down view at a specified instant $t$, the difference in distance $\Delta D(t) = D_B(t) - D_A(t)$ ($\Delta D > \SI{0}{m}$ means that agent $B$ is closer to the intersection), the agents' velocities $v$, their accumulated evidence (surprise) $E$, and the components of the pragmatic value associated with the current policy of each agent  (Equation~\eqref{eq:pragmatric_implement}, scaled with the surprise gain factor $\lambda$ from Equation~\eqref{eq:accumulated_surprise}). Future trajectories in the top-down view plots illustrate each agent's future states as predicted by another agent, with the corresponding state shown in the other plots as well. For instance, in panel \textbf{b} at $t=\SI{6.6}{s}$, agent $B$ predicts that agent $A$ would on average pursue a constant velocity (while in reality, agent $A$ decided to break shortly thereafter). The faint colored lines in the distance difference plots show the beliefs the agents have over $\Delta D$ (solid black lines show the actual outcome); for example, in panel \textbf{c}, agent $A$ overestimates $\Delta D$, while agent $B$ underestimates it. In the velocity plots, the faint colored lines indicate the beliefs about the velocity value held by the other agent (e.g., in panel \textbf{a} for $t>\SI{4.4}{s}$, agent $B$'s belief about agent $A$'s current velocity is consistently an overestimation).
    \textbf{a}) Early resolution: Agent $A$ starts $\SI{5}{m}$ closer to the intersection than agent $B$. When agent $B$ -- in response to the impending collision -- brakes first, agent $A$ becomes sufficiently confident that it can pass without having to also slow down; it crosses the intersection first without braking. Thus, the space-sharing conflict is here resolved by ``early yielding'' by agent $B$. 
    \textbf{b}) Late resolution: The two agents start at an equal distance to the intersection and initially start braking at the same time. However, agent $A$ brakes slightly harder, which gives agent $B$ confidence that it can go first (i.e., reducing uncertainty about the potential space overlap, opening up an affordance for $B$ to pass). This is then ``confirmed'' by $A$ through slowing down further. 
    \textbf{c}) Deadlock: Similar to panel (b), the two agents start at an equal distance to the intersection and start braking roughly at the same time. However, in this case the symmetry is never broken and the vehicles end up in a standstill where none of the agents are confident enough of a collision-free path to attempt passing, each believing that the other agent may start crossing at any moment.}
    \label{fig:no_norm_example}
\end{figure*}

Figure~\ref{fig:no_norm_example}\textbf{a} presents an example interaction where an agent with an initial lead ($\Delta D(0) = \SI{-5}{m}$) uses that lead to go first. Here, both agents initially predict a collision, as illustrated by low pragmatic values (high collision costs). For the trailing agent $B$, this is based on the assumption that the leading agent $A$ may slow down while the leading agent assumes that the trailing agent might accelerate. However, as the latter would violate the general normative expectation that drivers will not exceed the speed limit, such acceleration is treated as unlikely.  Therefore, the leading agent $A$ accumulates evidence for a full re-plan more slowly than the trailing agent $B$. Agent $B$ accumulates enough evidence for a re-plan at $t=\SI{3.8}{s}$ (first dashed line), selecting a policy that involves braking. Once the leading agent perceives this braking (second dashed line at $t = \SI{4.4}{s}$), it predicts an increasing lead (thin blue lines in the $\Delta D$ plot), thus perceiving an affordance to move ahead through the intersection with no collision risk. Consequently, its (negative) pragmatic value $G_{\text{prag}, A}$ approaches zero in all components, meaning there is little evidence for the need of a full re-plan; therefore, $A$ follows its original policy of crossing the intersection without braking. This represents a clear case where an initial difference in distance to the intersection is enough to yield an efficient resolution of the space-sharing conflict.

When the initial distances to the intersection are similar, stochastic variations in agents' policy generation might or might not break down the initial symmetry. For instance, in Figure~\ref{fig:no_norm_example}\textbf{b}, both agents start with an equal distance to the intersection, and therefore accumulate evidence for a full re-plan at a similar rate. They both initiate deceleration with the goal of letting the other vehicle pass. Here, a small difference in the agents' braking policies (due to stochasticity in policy selection) breaks the symmetry, leading to sufficiently different distances to the intersection ($\Delta D \not\approx 0$). The now trailing agent $A$ decides to slow down even more, which gives the leading agent $B$ enough confidence to cross the intersection first. 

However, if the initial braking policies of the agents happen to be similar (Figure~\ref{fig:no_norm_example}\textbf{c}), this leads to $\Delta D \approx 0$, with both agents continuing to brake until coming to a stop (dashed line at t $t=\SI{8.6}{s}$). In this situation, they end up in a deadlock, as there is no normative expectation that allows them to disregard the kinematically likely occurrence of the other vehicle accelerating onto the intersection. Specifically, this can be seen in the predictions of increasing velocities at $t = \SI{8.6}{s}$, which indicate that each agent perceives it as likely that the other agent will accelerate into the intersection (see the $\Delta D$ plot, where $A$ predicts $B$ taking the lead and vice versa). Thus, the space-sharing conflict is not resolved because implicit communication alone does not sufficiently reduce the agents' believed likelihood of a possible future collision.

\subsubsection{Normative expectations}\label{sec:norm_examples}

\begin{figure*}
    \centering
    \includegraphics[]{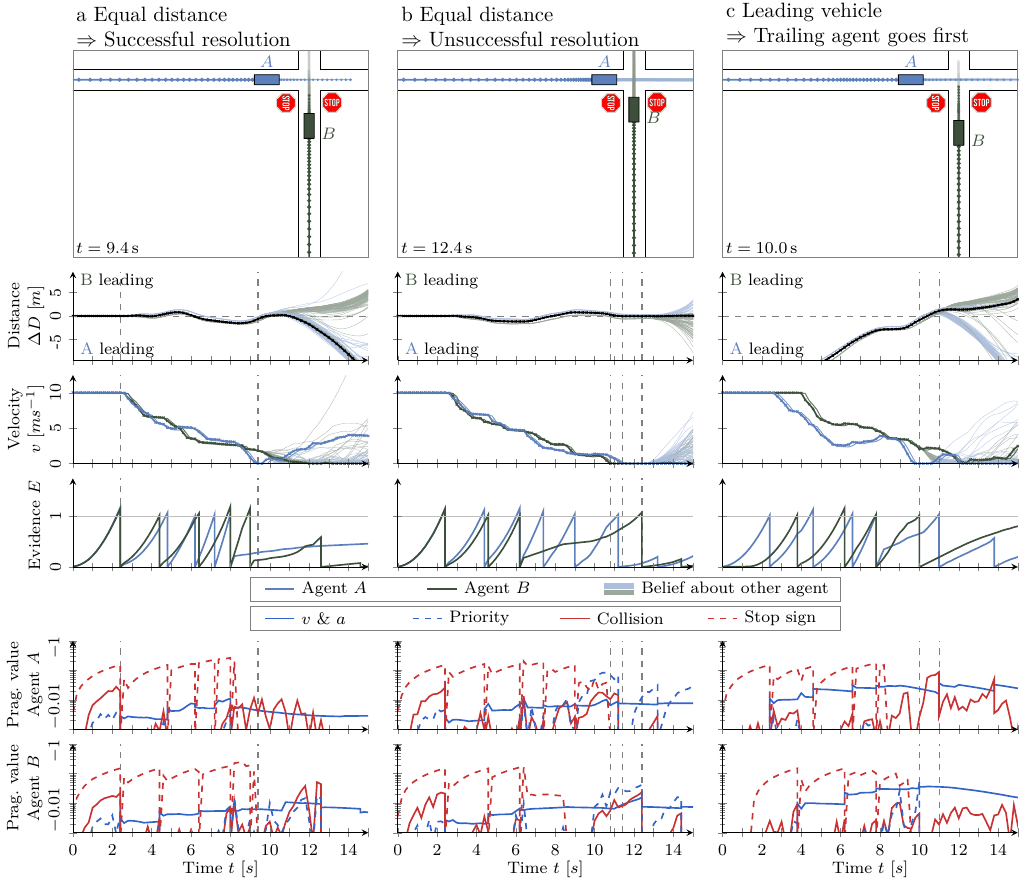}
    \caption{Three example interactions of the model endowed with normative expectations (priority rule and stop signs). \textbf{a}) The two agents start at an equal distance to the intersection, both agents consider the less-aggressive braking agent $A$ as the leading agent which allows $A$ (after stopping) to pass the intersection first. \textbf{b}) The two agents start at an equal distance to the intersection, but due to lagging perception, assign each other priority, leading to a standstill. \textbf{c}) The required stopping at intersections with priorities can lead to reversal in roles if the trailing agent $B$ decides to brake for the stop sign very late and aggressively. The reset of the accumulated evidence without reaching the threshold towards the end of the simulations is caused by automatic re-planning of the policy if an agent was standing still sufficiently long.}
    \label{fig:with_norm_example}
\end{figure*}
Our simulations demonstrated that the addition of interaction-specific norms can help resolve uncertainty in an interaction beyond what is possible with implicit communication (Figure~\ref{fig:prob_results}b). 
Figure~\ref{fig:with_norm_example}\textbf{a} shows an example simulation illustrating how this happens. Here, both agents perform initially similar braking maneuvers (from around $\SI{10}{m.s^{-1}}$ at $t = \SI{2.4}{s}$, first dashed line, to around $v = \SI{2}{m.s^{-1}}$). This -- as can be seen in the components of the pragmatic value -- is mostly driven by the need to follow the rules of the stop sign. Because of the normative expectation that other agents will brake for the stop sign, collisions are seen as much less likely (at $t = \SI{2.4}{s}$, the collision avoidance component of the pragmatic value is only $\approx -0.03$, while in the scenarios without normative expectation shown for instance in Figure~\ref{fig:no_norm_example}\textbf{b} it approaches $-0.1$). Therefore, the risk of collision is not the main driver of surprise accumulation here. However, even though agent $A$ continues to brake (coming to a complete stop at $t=\SI{9.4}{s}$, second dashed line), it is able to maintain a slight lead ($\Delta D < 0$). Consequently, it sees itself in the priority position. Combined with the fact that it assumes that agent $B$ still needs to brake until standstill as well (the thin blue lines for the predictive belief all trending toward negative $\Delta D$ after the second dashed line), agent $A$'s belief in a widening gap allows it to follow its plan of going first. 
This indicates that there are two mechanisms which facilitate the successful resolution of these initially symmetric scenarios. On the one hand, the stop signs impose a need for larger decelerations, which leaves more space for the stochasticity of the model to result in diverging policies, thereby breaking the symmetry. On the other hand, the assignment of priority allows agents to exploit smaller asymmetries (compared to the baseline case), as long as they are large enough to assign priority to oneself.

However, as discussed above, deadlock is still possible when interaction is guided by normative expectations, as illustrated in Figure~\ref{fig:with_norm_example}\textbf{b}. Here, the deadlock is mostly caused by the agents believing that the other one has the priority. As described in the model implementation section, the priority decision becomes final once an agent completes the deceleration in front of the stop sign. For agent $B$, this is the case at $t=\SI{10.8}{s}$, at which point it believes in agent $A$'s priority ($B$ believes that $\Delta D < 0$ and hence that agent $A$ has priority). Meanwhile, a short time later at $t=\SI{11.4}{s}$, agent $A$ sees agent $B$ in the lead (according to $A$'s belief,  $\Delta D > 0$). Consequently, both agents predict the other one to take the lead, with agent $B$ predicting a decrease in $\Delta D$ and agent $A$ predicting a further increase. This results in both agents standing still, with any advance considered too risky in terms of a potential collision.

As discussed above, under our implemented stop sign and priority norms, it is possible for an agent trailing by a large initial distance difference to still cross the intersection first (Figure~\ref{fig:prob_results}\textbf{b}). Figure~\ref{fig:with_norm_example}\textbf{c} illustrates how this happens: the trailing agent $B$ brakes so late and so harshly that at the point where it fulfills the speed requirement set by the stop signs, it is actually leading, and then goes first. This is highlighted by the negligible contribution of the priority component in agent $B$'s pragmatic value after its corresponding re-plan at $t = \SI{10.0}{s}$ (the first dashed line), as well as $B$'s confidence that agent $A$ will not increase speed (see the velocity plot). Similarly, as agent $A$ reached this point earlier while still leading, agent $A$ also considers itself as having priority (as can be seen in its prediction that the other agent will decelerate and therefore yield). However, as seen by the accelerating evidence accumulation and the decrease in the collision component of the pragmatic value after the first dashed line, agent $A$ no longer believes in the norm compliance of agent $B$, and therefore decides to brake instead at $t=\SI{11}{s}$ (second dashed line). While this scenario with both agents believing they have priority might induce collisions, we only observed collisions in $0.833\%$ of simulations in this scenario. While such behavior might seem as a model artifact, it has been shown that humans can interpret other traffic participants' cautious approach to interactions as yielding, and therefore self-assign priority and move aggressively~\cite{sucha2017pedestrian}. 
The same mechanism, where the leading agent brakes early while the trailing agent brakes late and hard, is likely the cause for the deadlocks observed in scenarios where one agent has a large initial lead (Figure~\ref{fig:prob_results}\textbf{b}).

\subsubsection{Explicit communication}\label{sec:communication_results} 

\begin{figure}
    \centering
    \includegraphics[]{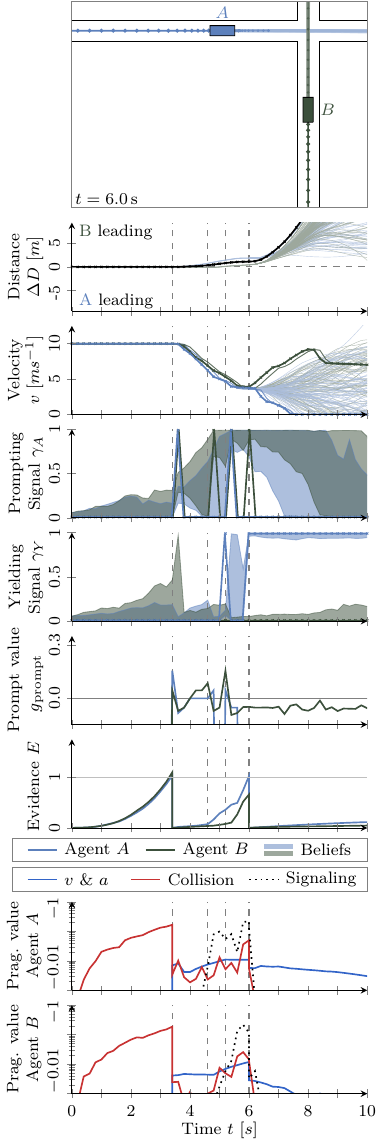}
    \caption{An example scenario for the interaction in which agents could use explicit communication. In addition to variables presented in previous examples, the explicit communication signals are shown (prompting $\gamma_{A}$ and yielding $\gamma_{Y}$ have slightly offset $x$-axis values for better visibility); shaded regions denote the range in the other agent's beliefs about each signal. Both agents start at equal distance to the intersection; agent $B$ is braking slightly less sharply than agent $A$ and is thus recognized by both agents as the leading vehicle. Both agents prompt each other for their intention (see $\gamma_A$), driven by the epistemic value of such an action (as defined in Equation~\eqref{eq:g_prompt}). After this exchange, agent $A$ signals yielding (see $\gamma_Y$), thereby reducing agent $B$'s  uncertainty in its belief about agent $A$. This gives agent $B$ the necessary confidence to cross the intersection first.}
    \label{fig:communication_example}
\end{figure}

We found that explicit communication in our model is highly efficient in enabling successful interactions (Figure~\ref{fig:prob_results}c). An example can be seen in Figure~\ref{fig:communication_example}, where two agents with initially equal distance to the intersection approach each other. After their first re-plan at $t = \SI{3.4}{s}$ (first dashed line), the agents initially begin to decelerate and concurrently prompt each other for their intention, driven by the predicted epistemic benefit of that action, $g_{\text{prompt}}$ which represents the difference between the epistemic value of prompting and the corresponding pragmatic cost (see Equation~\eqref{eq:g_prompt}). The dynamics of the prompting signals $\gamma_A$ however show that this initial attempt at communication is not yet fully perceived by the respective other agents. While for each agent the uncertainty about the other agent's yielding intention ($\gamma_Y$) is reduced initially by the lack of response to the prompting signal, it starts to build up again (see first the sharp decrease in $g_{\text{prompt}}$ after the initial prompting and then the gradual increase). By $t=\SI{4.6}{s}$ (second dashed line), the uncertainty in agent $B$'s belief has increased sufficiently for it to decide to prompt agent $A$ again at the next timestep. This repeated prompting is now sufficient for agent $A$ to become fully aware of agent $B$'s desire for a response, and given that it considers itself to be trailing, agent $A$ sends the first yielding signal $t=\SI{5.2}{s}$ (third dashed line). After a final prompt by agent $B$, and with the trailing margin increasing, agent $A$ commits to a continuous yielding signal at $t=\SI{6}{s}$ (fourth dashed line), which makes further prompting by agent $B$ unnecessary. Here, Agent $B$'s decision to signal yielding or not is fluctuating over time, as it based on a continuously updating evaluation of the kinematic scenario. Specifically, while it considers yielding for the first time at $t=\SI{5.2}{s}$ (third dashed line), it only commits to the yielding decision at $t=\SI{6}{s}$ (fourth dashed line).

Being convinced of agent $A$'s intention to yield, agent $B$ considers crossing the intersection to be safe (the gray lines in the $\Delta D$ plot indicate that agent $B$ predicts its small lead to increase significantly, while the drastic drop in the collision part of the pragmatic value shows that agent $B$ no longer considers collisions to be possible). Consequently, it decides to accelerate and cross the intersection first, avoiding a potential deadlock.

Not every instance of $g_{\text{prompt}} > 0$ was followed by a prompting signal ($\gamma_A = 1$), nor did every recorded prompting signal correspond to $g_{\text{prompt}} > 0$. This occurs because we did not save the predicted beliefs underlying $g_{\text{prompt}}$ (Equation~\eqref{eq:g_prompt}) during the simulation, recording only the current belief state and reconstructing $g_{\text{prompt}}$ \emph{post-hoc}. Noise in the state transition function can then lead to slightly different predicted beliefs and, consequently, different values of $g_{\text{prompt}}$ compared to the original simulation. Nevertheless, most prompting signals still coincide with the highest peaks in $g_{\text{prompt}}$, confirming that prompting actions are guided by their epistemic value.

\subsubsection{Explicit communication and normative expectations}
\begin{figure*}
    \centering
    \includegraphics[]{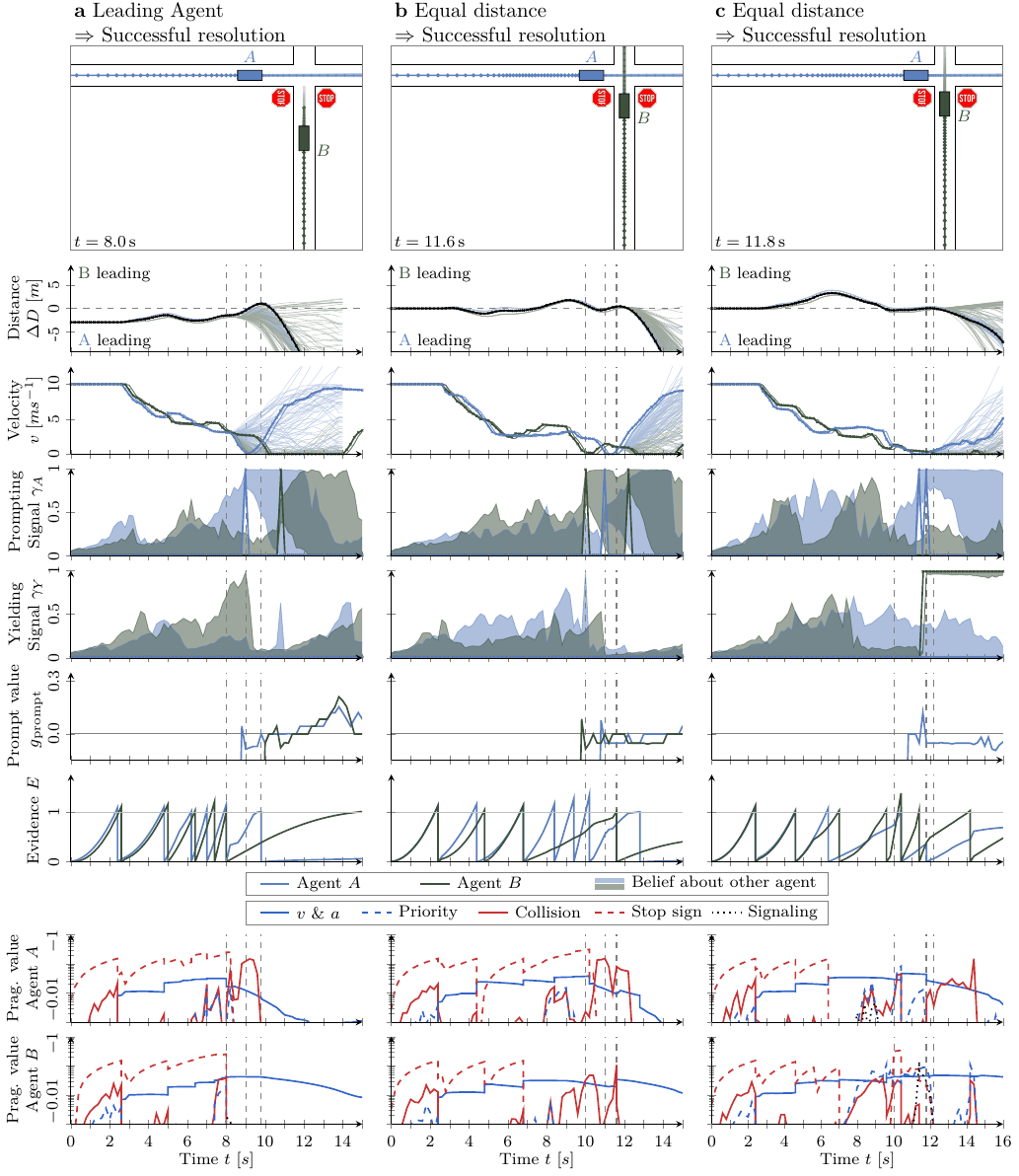}
    \caption{Three example interactions of the model equipped with explicit communication and normative expectations. \textbf{a}) At an intersection where agent $A$ is leading ($\Delta D (0) = \SI{-3}{m}$), it manages to pass first, solely relying on the normative expectation, with communication not necessary. \textbf{b}) The two agents start at an equal distance. As both agents assign themselves priority, they do not consider signaling yielding. In the end agent $A$ is faster at asserting its priority, allowing it to go first. \textbf{c}) Two agents again start at an equal distance to the intersection. Initially, both agents consider the other one to be leading and therefore having priority. However, with agent $B$ consistently signaling yielding, agent $A$ decides to cross the intersection.}
    \label{fig:communication_norm_example}
\end{figure*}

Similar to the case of uncontrolled intersections, in the presence of norms the capability for explicit communication improved the agents' likelihood to successfully resolve space-sharing conflicts (Figure~\ref{fig:prob_results}d). Examination of representative simulations showed that agents engaged in explicit communication only if proxemics and normative priors were insufficient to resolve the conflict. For instance, in Figure~\ref{fig:communication_norm_example}\textbf{a}, the initially leading agent $A$ ($\Delta D(0) = -\SI{3}{m}$) has a sufficient margin to execute the crossing policy decided upon at $t=\SI{8.0}{s}$ (first dashed line). Specifically, agent $A$ already deciding to cross first at $t=\SI{8.0}{s}$ is evident in two ways. First, by its next policy re-plan at $t=\SI{9.8}{s}$ (third dashed line), it is already accelerating, as seen in the velocity plot. Second, the thin blue trajectories in the $\Delta D$ plot indicate that, at $t=\SI{8.0}{s}$, agent $A$ predicts a substantial future lead, which is at that point only achievable by crossing the intersection.
Importantly, this confidence does not require explicit communication, as after receiving no response to its prompt at $t=\SI{9.0}{s}$ (second dashed line), agent $A$ assigns low probability to agent $B$’s (signaled) yielding intent $\gamma_Y$. Instead, agent $A$'s confidence is driven by its belief in its own priority, evident from the small priority norm component of the pragmatic value after $t=\SI{8.4}{s}$.

If both agents start at equal distances from the intersection, priority assignment might become more ambiguous. Nevertheless, in many such cases the agents can successfully resolve the conflict without having to rely on explicit signaling, with an example shown in Figure~\ref{fig:communication_norm_example}\textbf{b}. After stopping -- agent $B$ at $t=\SI{10}{s}$ (first dashed line) and agent $A$ at $t=\SI{11}{s}$ (second dashed line) -- both agents believe that they are leading (gray and blue curves in the $\Delta D$ plot at those respective timesteps). Consequently, each agent assigns itself priority, planning to cross first by accelerating right after coming to a stop (without a full policy re-plan). However, realizing that this would result in a collision, agent $B$ at $t=\SI{11.6}{s}$ decides to yield, allowing agent $A$ to assert itself and proceed onto the intersection first. In this example, agents did not rely on explicit communication because they assumed they could resolve the situation by relying on norms, so the pragmatic costs of communicative actions outweighed their epistemic benefits. Theoretically, the prompting in this scenario would be unnecessary, but with traffic norms not influencing the belief about yielding intent initially, some prompting is still observed.

However, there are other cases where communication is utilized extensively by the agents even in the presence of norms. For example, in Figure~\ref{fig:communication_norm_example}\textbf{c}, both agents assign priority to the other one -- contrasting Figure~\ref{fig:with_norm_example}b -- with both assuming the other agent to hold at the moment they come to a stop (agent $A$ at $t=\SI{10.0}{s}$ and agent $B$ at $t=\SI{12.2}{s}$ -- the first and third dashed lines, respectively). Without communication, this scenario would have most likely resulted in a deadlock (as in Figure~\ref{fig:with_norm_example}\textbf{b}). However, prompted by agent $A$, agent $B$ decides to communicate its intent by signaling yielding. Consequently, having perceived this signal at $t=\SI{11.8}{s}$ (third dashed line), agent $A$ predicts agent $B$ to keep standing still (thin gray lines in the velocity plot), resulting in high confidence of agent $A$ that crossing first is safe.

\subsection{Conflict from false signaling}
\begin{figure}
    \centering
    \includegraphics[]{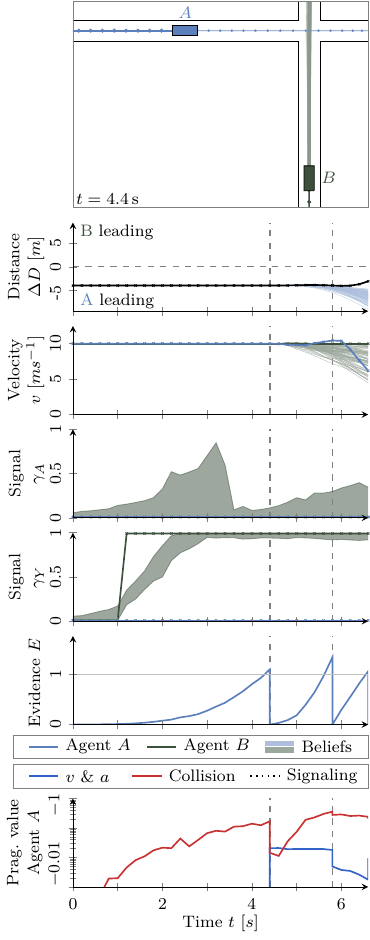}
    \caption{Simulation of model behavior in a provoked conflict. The leading agent $A$ ($\Delta D (0) = \SI{-4}{m}$) is modeled, while agent $B$ is programmed to drive with constant speed while signaling yielding, successfully deceiving agent $A$, which results in a collision.}
    \label{fig:critical_example}
\end{figure}
\begin{figure}
    \centering
    \includegraphics[]{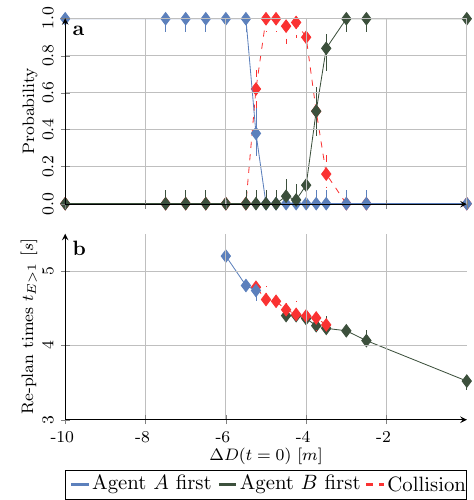}
    \caption{\textbf{a}) The likelihoods of the different simulation outcomes in the provoked conflict scenario, with error bars indicating 95\% confidence intervals.
    \textbf{b}) The time $t_{E>1}$ at which the agent $A$ first triggered a full re-plan of its policy (lower panel). Error bars show the 5th and 95th percentiles.}
    \label{fig:conflict_results}
\end{figure}
The above scenarios illustrate the interactive behavior of the agents under the assumption that both agents are fully attentive and cooperative; hence, the interaction is virtually never safety-critical. To investigate the model behavior in an adversarial setting, we pre-programmed one of the agents ($B$) to drive with a constant velocity while signaling yielding to mislead the modeled agent $A$ into causing a collision, with initial distances $\Delta D (0) \in [\SI{-10}{m}, \SI{0}{m}]$. Here, the yielding signal gives the modeled agent the belief of being able to pass safely, which is then violated by agent $B$.

Figure~\ref{fig:critical_example} presents an example of such interaction. There, at the first re-plan ($t=\SI{4.4}{s}$, first dashed line), convinced by the yielding signal of the other vehicle, the leading agent $A$ considers light acceleration to be sufficient to cross the intersection safely. The collision part of the pragmatic values falls by over an order of magnitude, and $A$'s predictions of $\Delta D$ show high confidence that the gap will become wider, with agent $B$ predicted to brake in the velocity plot. However, over the following timesteps, agent $A$ slowly loses trust into the other agent's signaling, with collisions being considered increasingly likely under the current policy (the collision component of the pragmatic value increases). Specifically, not observing the increasingly larger (and, according to the state transition function, increasingly improbable) decelerations required from agent $B$ to still yield safely, agent $A$ no longer expects $B$ to yield. This finally convinces $A$ that braking is needed (this is done through the second full policy re-plan at $t = \SI{5.8}{s}$, the second dashed line). However, at this point, a collision is no longer avoidable, and braking mainly serves the purpose of minimizing the relative impact velocity.

This pattern of the modeled agent $A$ being unable to avoid a collision seems to be mostly limited to certain values of $\Delta D(0)$, ranging roughly in between $\SI{-5.5}{m}$ and $\SI{-3.5}{m}$ (Figure~\ref{fig:conflict_results}\textbf{a}). For larger initial leads, the agent is able to simply go first by maintaining its velocity or slightly accelerating, while for much smaller initial leads ($\Delta D(0) \geq \SI{-3.5}{m}$), it is able to brake in time. Specifically, the ability of the agent to avoid collisions seems to mostly depend on it being able to respond earlier, as shown in Figure~\ref{fig:conflict_results}\textbf{b}. Given how the initial response time is driven purely by predicted collisions (see pragmatic value in Figure~\ref{fig:critical_example}), shorter response times are caused by the agent believing in the possibility of a collision earlier, which is conditioned on them losing trust in the other agent's signaled intent. Consequently, with an earlier reaction and less reliance on the agent $B$'s yielding signal, agent $A$ is able to choose a policy which avoids a collision.

\section{Discussion}
In this paper, we extended an existing active inference-based computational driver model~\cite{engstrom_resolving_2024,schumann_active_2025} to account for interactive behavior of two road users, mirroring a novel recent framing of social cognition in the active inference literature~\cite{friston2015active, constant2019regimes, pezzulo_predictive_2026, vasil_world_2020, constant2018variational, bruineberg2018free, lee2026editorial}. In this framing, the interacting agents form mutual and reciprocal beliefs about each others' future actions. If the agents' mutual beliefs are well aligned, a common ground is established and the interactive situation can be seen as governed by a single agent-neutral generative model, like when two persons are singing or dancing together. However, such alignment may be precluded by high initial uncertainty in the agents' beliefs, which can be resolved by two key means: (1) reliance on normative expectations and (2) explicit communication.

Our simulation results demonstrate how our model can provide a concrete mechanistic account of this general framing of road user interaction. In the baseline simulations (Figures~\ref{fig:prob_results}\textbf{a}~and~\ref{fig:no_norm_example}), without interaction-specific normative expectations, the space-sharing conflicts can still be resolved by implicit communication as long as the initial distances between the agents are large enough (at least \SI{5}{m}, corresponding to a \SI{0.5}{s} time gap). In such situations, the trailing agent typically starts yielding which reinforces the leading agent’s policy to proceed first as it, despite the uncertainty, perceives that it is safe to move ahead.  
However, this resolution mechanism breaks down for smaller initial distances where the mutual uncertainty about the other agent’s future behavior often leads to deadlock.

Adding interaction-specific normative expectations -- the ``first come, first go'' priority rule supported by a stop sign (a deontic cue) requiring agents to come to a stop before proceeding -- constrains the beliefs of the agents and usually leads to a successful resolution (Figures~\ref{fig:prob_results}\textbf{b}~and~\ref{fig:with_norm_example}). This mechanism, however, typically fails in situations where the agents apply the rule in a non-congruent way (\emph{e.g.}, both agents believe that the other agent arrived first and thus has priority).

Our simulations also show how explicit communication (implemented by simple communicative acts such as signaling yielding in response to initial prompting), offers an efficient way to resolve uncertainty in situations that are not guided by clear normative expectations (Figures~\ref{fig:prob_results}\textbf{b}~and~\ref{fig:with_norm_example}), or when the agents' interpretations of who has priority are not aligned (Figure~\ref{fig:communication_norm_example}\textbf{c}).
Finally, we demonstrate how collisions can occur as the result of false expectations, in this case overly confident beliefs in misleading signaling by the other agent (Figures~\ref{fig:critical_example}~and~\ref{fig:conflict_results}). 

Our proposed framing and computational model connect road user interaction to broader contemporary accounts of social cognition, in particular the classical notion of common ground in linguistic communication~\cite{clark1991grounding} and modern, active inference-based approaches to joint action based on shared, agent-neutral, generative models~\cite{friston2015active, pezzulo_predictive_2026}.  
In particular, we leverage the idea that successful road user interaction requires that agents establish a common understanding of the situation and, as a consequence, can be seen as governed by a shared generative model. In our current approach, each agent retains their own generative model which generates possible futures, including their own planned actions as well as the predicted actions of the other agent. If these possible futures are aligned, the agents thus have established a single common expectation of how the situation will play out and the agents can be seen as being governed by a single agent-neutral model, as suggested by Pezzulo \emph{et al.}~\cite{pezzulo_predictive_2026} for other forms of joint action. However, in our framing, such an agent-neutral model should not be seen as separate from the agents’ individual generative models, but rather as a phenomenon that is established when two agents enter into a coupled interaction. 

Our model also highlights how road user interaction is facilitated by normative expectations and explicit communication as different means for reducing uncertainty in order to establish a shared and precise understanding (generative model) of the traffic situation. 
Here, normative expectations, facilitated by deontic cues such as road signs, traffic lights and lane markings, can be understood as an example of niche construction~\cite{constant2018variational,bruineberg2018free} as part of a general \emph{traffic culture} that differs between geographical and cultural regions. On the other hand, explicit communication (here prompting the other agent for its intent to yield) can be seen as a specific form of epistemic action (driven by epistemic value) with the purpose to gain uncertainty-resolving information, thus serving a general function similar to that of visually scanning an intersection or checking the rear view mirror before changing lane~\cite{engstrom_resolving_2024}. 

Our model and simulations suggest an explanation for the existing empirical finding that close dynamical coupling in road user interactions is primarily observed in uncontrolled scenarios~\cite{noonan2022interdependence}. Such scenarios are similar to the baseline uncontrolled intersection in the present simulations (Figure~\ref{fig:with_norm_example}), where the decision on who goes first and who yields emerges dynamically based on the scenario kinematics, leading to a basic form of coupling. In this coupling, once the symmetry is broken, both vehicles will act jointly to reinforce the resolution of the space-sharing conflict such that their actions become interdependent (one vehicle will continue at the current speed or accelerate and the other will decelerate to yield). However, in situations where the agents rely mainly on normative expectations (\emph{e.g.}, in our stop sign + precedence rule scenario; Figure~\ref{fig:with_norm_example}), the agents' behavior will be less interdependent at the sensorimotor level.

The model also accounts for the well-established finding that explicit communication is relatively rare in real-world road user interactions~\cite{lee2021road}. Since the space-sharing conflict can most often be resolved by implicit communication and direct coupling (as in Figure~\ref{fig:no_norm_example}) or by relying on established normative expectations (Figure~\ref{fig:with_norm_example}), explicit communication only becomes necessary when any of these mechanisms fail, for example when mutual uncertainty about each others' future action prevents progress (as in Figures~\ref{fig:no_norm_example}\textbf{c}) or when normative expectations are misinterpreted (as in Figure~\ref{fig:communication_norm_example}\textbf{c}).    

Compared to existing road user interaction models~\cite{camara2018towards,camara_pedestrian_2021,fox2018should,fisac2019hierarchical,kang2017game, schwarting_social_2019, thalya2020modeling, sadigh_planning_2018, boda2020computational, giles2019zebra, markkula_evidence_2018, pekkanen_variable-drift_2022, zgonnikov_should_2024, domeyer2020interdependence,noonan2022interdependence, johora2020zone, siebinga_model_2024-1} our model adds several novel features similar in spirit to recent computational models of social cognition and behavior outside the traffic domain~\cite{friston2015active, constant2019regimes, pezzulo_predictive_2026, vasil_world_2020, constant2018variational, bruineberg2018free}. First, it proposes distinct mechanistic roles for dynamical coupling (\emph{i.e.}, implicit communication), normative expectations and explicit communication in the resolution of space-sharing conflicts. Second, in contrast to most existing models, our model posits the resolution of uncertainty in the agent's mutual beliefs about each other as the key factor underlying road user interaction behavior. Finally, many existing models are based on the game-theoretic notion of inferring and reasoning about the intent of the other agents and one's own actions in a nested and explicit way (\emph{e.g.}, ``if I do $a$, they will do $b$, and then I can do $c$, etc.''). While this type of explicit reasoning is certainly possible, it does not seem necessary and may play a less important role in road user interaction than traditionally believed (see Pezzulo~\emph{et al.}~\cite{pezzulo_predictive_2026}  for a similar argument in the context of joint action).

While the model presented here provides a novel mechanistic account of several key aspects of road user interaction, it is necessarily simplified in many ways and subject to a number of limitations.
First, in the current model, the reduction of uncertainty in beliefs is essentially ``built into'' the model rather than arising from an explicit inference process. As a result, the uncertainty in the explicit belief about the other agent's yielding intent can only be reduced through explicit communication, while implicit communication (\emph{e.g.}, kinematics) and contextual information such as traffic rules do not directly update that belief. This limitation can lead to implausible behavior. For example, an agent’s belief about the other agent’s yielding intent initially evolves regardless of the agents' proxemics, at least until communication begins. Consequently, an agent may be equally likely to prompt the other for its intention regardless of whether both vehicles arrive at the intersection at nearly the same time or one arrives substantially earlier. In the latter case, however, such prompting is unrealistic, since the probability that the trailing agent proceeds first is negligible -- there is little actual uncertainty about the outcome. A more principled approach would allow beliefs about yielding intent to be updated through a richer Bayesian inference process that incorporates kinematic information, traffic rules, and potentially implicit communicative cues. Under such a formulation, a substantial lead by one agent would already reduce uncertainty about yielding intent, rendering the epistemic value of prompting negligible compared to the pragmatic cost of that action (see also Maisto \emph{et al.}~\cite{maisto2023interactive} for a related approach). However, such an approach would require careful balancing to avoid positive feedback: high confidence in the other agent yielding would bias an agent towards predicting kinematics that reflect yielding behavior, which in turn could provide confirmatory evidence that further reinforces this belief—without any external evidence justifying it.

Second, the model does not account for accidental misunderstandings, and unrealistically assumes that signal observability is independent of distance or relative velocity. Specifically, given reduced visibility through for example car windshields, it would be reasonable to assume that perceiving a human driver's signal would require a close spatial proximity. Additionally, for such a mechanism to work properly, one would also have to imbue the modeled driver with an understanding that its own signals might not always be perceived. Furthermore, the exact method for the discussed signals (prompting and yielding) has been ignored for simplicities sake.

Third, one main limitation with the current model is that the implemented mechanism underlying the perception and selection of communicative acts is not fully integrated with the reasoning about kinematic policies. As a result, the model is of a somewhat \emph{ad hoc} nature; for example, kinematic policies are re-planned upon perceiving a yielding signal independently of the evidence accumulation and policy selection processes. Consequently, the implementation of explicit communication should be regarded as a proof-of-concept illustrating the general idea that communication between road users can be framed in terms of active inference principles, and more specifically, as being driven by epistemic value. To obtain a more realistic communication model, further work is needed to better integrate the selection of signal actions into the existing kinematic policy selection model~\cite{schumann_active_2025}. We view this as a key area of future research.

Finally, we did not compare our model to human data. Such data is not readily available for the interactive scenario analyzed here; our results were consistent with the more general observations reported in the literature (\emph{e.g.}, the prevalence of implicit communication in human road user interactions~\cite{lee_road_2021}), but future work should focus on bringing the model closer to scenarios for which detailed human data are available or could be collected. Still, even without direct comparisons to human data, our results are valuable, demonstrating that active inference can unify communication, coupling, and normative mechanisms into a general framework of road user interaction while producing results consistent with general intuition. Additionally, our model can explain why communication and norms are needed to increase the efficiency of traffic interactions, while highlighting the danger that adversarial agents can pose (for example by misleading communication). Furthermore, by bringing the model to an interactive scenario, we expand its potential usefulness in the simulation-based evaluation of automated vehicles~\cite{fremont2020formal,montali_waymo_2023,uzzaman2025testing}.
Importantly, our insights about the importance of uncertainty reduction (and the role expectations and communication play in it) could be transferred to other human interactions as well, such as pedestrians on sidewalks~\cite{siebinga2025model}.

\subsection{Conclusion}
In this work, we used a novel model, based on active inference and implementing communication and normative expectations, to simulate the interaction between two agents at an intersection. We showed that both communication and traffic norms can be used to increase the likelihood of a successful and timely interaction by reducing the uncertainty in road users' beliefs about each other. However, this depends on the agents being collaborative and acting in good faith, and we show that misleading intention signaling in road user interactions can cause collisions. Our results, although not validated against human data, demonstrate that active inference can provide a unified account of the role of behavioral coupling, communication, and normative mechanisms in road user interaction. Additionally, we highlight the risks introduced by adversarial or misleading agents.

\begin{dci}
    JE, RW and SYL were employed by Waymo LLC and conducted the research without any external funding from third-parties. Techniques discussed in this paper may be described in U.S. Patent Application Nos. 18/614,428 and 63/657,623. 
\end{dci}

\begin{funding}
    Delft University of Technology received funding from Waymo LLC for parts of the research carried out by JFS, but JFS received no direct financial benefit for his contributions to the paper.
\end{funding}
\bibliographystyle{SageV}
\bibliography{zotero_reference, manual_bib}

@article{abbas_drivers_2024,
  title={Drivers’ Knowledge, Attitude and Practices Towards Traffic Rules and Regulations},
  author={Abbas, Sheer and Fatima, Sidra and Sharif, Azhar and Adnan, Shiekh Muhammad},
  journal={Journal of Road Safety},
  volume={35},
  number={3},
  pages={24--31},
  year={2024},
  publisher={Australasian College of Road Safety}
}

@article{domeyer2022driver,
  title={Driver-pedestrian perceptual models demonstrate coupling: Implications for vehicle automation},
  author={Domeyer, Joshua E and Lee, John D and Toyoda, Heishiro and Mehler, Bruce and Reimer, Bryan},
  journal={IEEE Transactions on Human-Machine Systems},
  volume={52},
  number={4},
  pages={557--566},
  year={2022},
  publisher={IEEE}
}

@article{amundsen1977proceedings,
  title={Proceedings of first workshop on traffic conflicts},
  author={Amundsen, FH and Hyden, C},
  journal={Oslo, TTI, Oslo, Norway and LTH Lund, Sweden},
  volume={78},
  year={1977}
}

@article{maisto2023interactive,
  title={Interactive inference: A multi-agent model of cooperative joint actions},
  author={Maisto, Domenico and Donnarumma, Francesco and Pezzulo, Giovanni},
  journal={IEEE Transactions on Systems, Man, and Cybernetics: Systems},
  volume={54},
  number={2},
  pages={704--715},
  year={2023},
  publisher={IEEE}
}

@inproceedings{dey2017pedestrian,
  title={Pedestrian interaction with vehicles: roles of explicit and implicit communication},
  author={Dey, Debargha and Terken, Jacques},
  booktitle={Proceedings of the 9th international conference on automotive user interfaces and interactive vehicular applications},
  pages={109--113},
  year={2017}
}

@article{zgonnikov2024nudging,
  title={Nudging human drivers via implicit communication by automated vehicles: Empirical evidence and computational cognitive modeling},
  author={Zgonnikov, Arkady and Beckers, Niek and George, Ashwin and Abbink, David and Jonker, Catholijn},
  journal={International Journal of Human-Computer Studies},
  volume={185},
  pages={103224},
  year={2024},
  publisher={Elsevier}
}

@article{siebinga2023modelling,
  title={Modelling communication-enabled traffic interactions},
  author={Siebinga, Olger and Zgonnikov, Arkady and Abbink, David A},
  journal={Royal Society open science},
  volume={10},
  number={5},
  year={2023},
  publisher={The Royal Society}
}

@misc{lee2026editorial,
  title={Editorial to the Special Issue on Road User Interactions in the Age of Vehicle Automation},
  author={Lee, Yee Mun and Domeyer, Joshua},
  journal={Transportation Research Part F: Traffic Psychology and Behaviour},
  volume={116},
  pages={103439},
  year={2026},
  publisher={Elsevier}
}

@article{miller2022implicit,
  title={Implicit intention communication as a design opportunity for automated vehicles: Understanding drivers’ interpretation of vehicle trajectory at narrow passages},
  author={Miller, Linda and Leitner, Jasmin and Kraus, Johannes and Baumann, Martin},
  journal={Accident Analysis \& Prevention},
  volume={173},
  pages={106691},
  year={2022},
  publisher={Elsevier}
}

@article{tian2023deceleration,
  title={Deceleration parameters as implicit communication signals for pedestrians’ crossing decisions and estimations of automated vehicle behaviour},
  author={Tian, Kai and Tzigieras, Athanasios and Wei, Chongfeng and Lee, Yee Mun and Holmes, Christopher and Leonetti, Matteo and Merat, Natasha and Romano, Richard and Markkula, Gustav},
  journal={Accident Analysis \& Prevention},
  volume={190},
  pages={107173},
  year={2023},
  publisher={Elsevier}
}

@article{bruineberg2018free,
  title={Free-energy minimization in joint agent-environment systems: A niche construction perspective},
  author={Bruineberg, Jelle and Rietveld, Erik and Parr, Thomas and van Maanen, Leendert and Friston, Karl J},
  journal={Journal of theoretical biology},
  volume={455},
  pages={161--178},
  year={2018},
  publisher={Elsevier}
}

@article{constant2018variational,
  title={A variational approach to niche construction},
  author={Constant, Axel and Ramstead, Maxwell JD and Veissiere, Samuel PL and Campbell, John O and Friston, Karl J},
  journal={Journal of the Royal Society Interface},
  volume={15},
  number={141},
  year={2018},
  publisher={The Royal Society}
}

@inproceedings{siebinga2025model,
  title={A model of the sidewalk salsa},
  author={Siebinga, Olger},
  booktitle={2025 IEEE International Conference on Systems, Man, and Cybernetics (SMC)},
  pages={7298--7305},
  year={2025},
  organization={IEEE}
}

@inproceedings{fremont2020formal,
  title={Formal scenario-based testing of autonomous vehicles: From simulation to the real world},
  author={Fremont, Daniel J and Kim, Edward and Pant, Yash Vardhan and Seshia, Sanjit A and Acharya, Atul and Bruso, Xantha and Wells, Paul and Lemke, Steve and Lu, Qiang and Mehta, Shalin},
  booktitle={2020 IEEE 23rd International Conference on Intelligent Transportation Systems (ITSC)},
  pages={1--8},
  year={2020},
  organization={IEEE}
}

@article{uzzaman2025testing,
  title={Testing Autonomous Vehicles in Virtual Environments: A Review of Simulation Tools and Techniques},
  author={Uzzaman, Asif and Islam, Monirul and Hossain, Md Shimul},
  journal={Control Syst. Optim. Lett},
  volume={3},
  pages={151--157},
  year={2025}
}

@article{wilson1927probable,
  title={Probable inference, the law of succession, and statistical inference},
  author={Wilson, Edwin B},
  journal={Journal of the American Statistical Association},
  volume={22},
  number={158},
  pages={209--212},
  year={1927},
  publisher={Taylor \& Francis}
}

@article{clark1991grounding,
  title={Grounding in communication.},
  author={Clark, Herbert H and Brennan, Susan E},
  year={1991},
  publisher={American Psychological Association},
  journal={Perspectives on socially shared cognition},
}

@article{sucha2017pedestrian,
  title={Pedestrian-driver communication and decision strategies at marked crossings},
  author={Sucha, Matus and Dostal, Daniel and Risser, Ralf},
  journal={Accident Analysis \& Prevention},
  volume={102},
  pages={41--50},
  year={2017},
  publisher={Elsevier}
}

@article{lee2021road,
  title={Road users rarely use explicit communication when interacting in today’s traffic: implications for automated vehicles},
  author={Lee, Yee Mun and Madigan, Ruth and Giles, Oscar and Garach-Morcillo, Laura and Markkula, Gustav and Fox, Charles and Camara, Fanta and Rothmueller, Markus and Vendelbo-Larsen, Signe Alexandra and Rasmussen, Pernille Holm and others},
  journal={Cognition, Technology \& Work},
  volume={23},
  number={2},
  pages={367--380},
  year={2021},
  publisher={Springer}
}

@article{constant2019regimes,
  title={Regimes of expectations: An active inference model of social conformity and human decision making},
  author={Constant, Axel and Ramstead, Maxwell JD and Veissi{\`e}re, Samuel PL and Friston, Karl},
  journal={Frontiers in psychology},
  volume={10},
  pages={679},
  year={2019},
  publisher={Frontiers Media SA}
}

@book{goffman2017interaction,
  title={Interaction ritual: Essays in face-to-face behavior},
  author={Goffman, Erving},
  year={2017},
  publisher={Routledge}
}

@article{friston2015active,
  title={Active inference, communication and hermeneutics},
  author={Friston, Karl J and Frith, Christopher D},
  journal={cortex},
  volume={68},
  pages={129--143},
  year={2015},
  publisher={Elsevier}
}

@article{noonan2022interdependence,
  title={Interdependence of driver and pedestrian behavior in naturalistic roadway negotiations},
  author={Noonan, T Zach and Gershon, Pnina and Domeyer, Josh and Mehler, Bruce and Reimer, Bryan},
  journal={Traffic injury prevention},
  volume={23},
  number={sup1},
  pages={S62--S67},
  year={2022},
  publisher={Taylor \& Francis}
}

@article{domeyer2020vehicle,
  title={Vehicle automation--other road user communication and coordination: Theory and mechanisms},
  author={Domeyer, Joshua E and Lee, John D and Toyoda, Heishiro},
  journal={IEEE Access},
  volume={8},
  pages={19860--19872},
  year={2020},
  publisher={IEEE}
}

@article{de2019external,
  title={External human-machine interfaces on automated vehicles: Effects on pedestrian crossing decisions},
  author={De Clercq, Koen and Dietrich, Andre and N{\'u}{\~n}ez Velasco, Juan Pablo and De Winter, Joost and Happee, Riender},
  journal={Human factors},
  volume={61},
  number={8},
  pages={1353--1370},
  year={2019},
  publisher={SAGE Publications Sage CA: Los Angeles, CA}
}

@article{domeyer2019proxemics,
  title={Proxemics and kinesics in automated vehicle--pedestrian communication: Representing ethnographic observations},
  author={Domeyer, Joshua and Dinparastdjadid, Azadeh and Lee, John D and Douglas, Grace and Alsaid, Areen and Price, Morgan},
  journal={Transportation research record},
  volume={2673},
  number={10},
  pages={70--81},
  year={2019},
  publisher={SAGE Publications Sage CA: Los Angeles, CA}
}

@article{portouli2014drivers,
  title={Drivers' communicative interactions: on-road observations and modelling for integration in future automation systems},
  author={Portouli, Evangelia and Nathanael, Dimitris and Marmaras, Nicolas},
  journal={Ergonomics},
  volume={57},
  number={12},
  pages={1795--1805},
  year={2014},
  publisher={Taylor \& Francis}
}

@article{domeyer2020interdependence,
  title={Interdependence in vehicle-pedestrian encounters and its implications for vehicle automation},
  author={Domeyer, Joshua E and Lee, John D and Toyoda, Heishiro and Mehler, Bruce and Reimer, Bryan},
  journal={IEEE transactions on intelligent transportation systems},
  volume={23},
  number={5},
  pages={4122--4134},
  year={2020},
  publisher={IEEE}
}

@article{habibovic2013driver,
  title={Driver behavior in car-to-pedestrian incidents: An application of the Driving Reliability and Error Analysis Method (DREAM)},
  author={Habibovic, Azra and Tivesten, Emma and Uchida, Nobuyuki and B{\"a}rgman, Jonas and Aust, Mikael Ljung},
  journal={Accident Analysis \& Prevention},
  volume={50},
  pages={554--565},
  year={2013},
  publisher={Elsevier}
}

@inproceedings{camara2018towards,
  year={2018},
  booktitle={IEEE International Workshop on Intelligent Robots and Systems (IROS)},
  title={Towards pedestrian-AV interaction: method for elucidating pedestrian preferences},
  author={Camara, Fanta and Cosar, Serhan and Bellotto, Nicola and Merat, Natasha and Fox, Charles W}
}

@article{fox2018should,
  title={When should the chicken cross the road},
  author={Fox, Charles and Camara, Fanta and Markkula, Gustav and Romano, Richard and Madigan, Ruth and Merat, Natasha and others},
  journal={Game theory for autonomous vehicle-human interactions},
  year={2018}
}

@inproceedings{fisac2019hierarchical,
  title={Hierarchical game-theoretic planning for autonomous vehicles},
  author={Fisac, Jaime F and Bronstein, Eli and Stefansson, Elis and Sadigh, Dorsa and Sastry, S Shankar and Dragan, Anca D},
  booktitle={2019 International conference on robotics and automation (ICRA)},
  pages={9590--9596},
  year={2019},
  organization={IEEE}
}

@article{kang2017game,
  title={Game theoretical approach to model decision making for merging maneuvers at freeway on-ramps},
  author={Kang, Kyungwon and Rakha, Hesham A},
  journal={Transportation Research Record},
  volume={2623},
  number={1},
  pages={19--28},
  year={2017},
  publisher={SAGE Publications Sage CA: Los Angeles, CA}
}

@inproceedings{thalya2020modeling,
  title={Modeling driver behavior in interactions with other road users},
  author={Thalya, Prateek and Kovaceva, Jordanka and Knauss, Alessia and Lubbe, Nils and Dozza, Marco},
  booktitle={Proceedings of 8th Transport Research Arena},
  year={2020},
}

@article{boda2020computational,
  title={A computational driver model to predict driver control at unsignalised intersections},
  author={Boda, Christian-Nils and Lehtonen, Esko and Dozza, Marco},
  journal={IEEE Access},
  volume={8},
  pages={104619--104631},
  year={2020},
  publisher={IEEE}
}

@inproceedings{giles2019zebra,
  title={At the zebra crossing: Modelling complex decision processes with variable-drift diffusion models},
  author={Giles, Oscar and Markkula, Gustav and Pekkanen, Jami and Yokota, Naoki and Matsunaga, Naoto and Merat, Natasha and Daimon, Tatsuru},
  booktitle={Proceedings of the Annual Meeting of the Cognitive Science Society},
  volume={41},
  year={2019}
}

@article{johora2020zone,
  title={Zone-specific interaction modeling of pedestrians and cars in shared spaces},
  author={Johora, Fatema T and M{\"u}ller, J{\"o}rg P},
  journal={Transportation research procedia},
  volume={47},
  pages={251--258},
  year={2020},
  publisher={Elsevier}
}

@article{friston2015duet,
  title={A duet for one},
  author={Friston, Karl and Frith, Christopher},
  journal={Consciousness and cognition},
  volume={36},
  pages={390--405},
  year={2015},
  publisher={Elsevier}
}

@book{bicchieri2016norms,
  title={Norms in the wild: How to diagnose, measure, and change social norms},
  author={Bicchieri, Cristina},
  year={2016},
  publisher={Oxford University Press}
}

@article{fraade2025being,
  title={Being good (at driving): Characterizing behavioral expectations on automated and human driven vehicles},
  author={Fraade-Blanar, Laura and Favar{\`o}, Francesca and Engström, Johan and Cefkin, Melissa and Best, Ryan and Lee, John and Victor, Trent},
  journal={arXiv preprint arXiv:2502.08121},
  year={2025}
}

@misc{schumann_active_2025,
    title = {Active inference as a unified model of collision avoidance behavior in human drivers},
    url = {http://arxiv.org/abs/2506.02215},
    doi = {10.48550/arXiv.2506.02215},
    urldate = {2025-07-23},
    publisher = {arXiv},
    author = {Schumann, Julian F. and Engström, Johan and Johnson, Leif and O'Kelly, Matthew and Messias, Joao and Kober, Jens and Zgonnikov, Arkady},
    month = jun,
    year = {2025},
    note = {arXiv:2506.02215 [cs]},
}

@misc{pezzulo_predictive_2026,
	title = {A {Predictive} {Processing} {Framework} for {Joint} {Action} and {Communication}},
	url = {https://osf.io/preprints/psyarxiv/q4jnr_v4/},
	urldate = {2026-02-26},
	publisher = {PsyArXiv},
	author = {Pezzulo, Giovanni and Knoblich, Günther and Maisto, Domenico and Donnarumma, Francesco and Pacherie, Elisabeth and Hasson, Uri},
	month = feb,
	year = {2026},
}

@article{zgonnikov_should_2024,
	title = {Should {I} {Stay} or {Should} {I} {Go}? {Cognitive} {Modeling} of {Left}-{Turn} {Gap} {Acceptance} {Decisions} in {Human} {Drivers}},
	issn = {0018-7208},
	shorttitle = {Should {I} {Stay} or {Should} {I} {Go}?},
	url = {https://doi.org/10.1177/00187208221144561},
	doi = {10.1177/00187208221144561},
	language = {en},
	urldate = {2023-02-08},
	journal = {Human Factors},
	author = {Zgonnikov, Arkady and Abbink, David and Markkula, Gustav},
	year = {2024},
	note = {Publisher: SAGE Publications Inc},
	pages = {00187208221144561},
}

@article{engstrom_resolving_2024,
	title = {Resolving uncertainty on the fly: modeling adaptive driving behavior as active inference},
	volume = {18},
	issn = {1662-5218},
	shorttitle = {Resolving uncertainty on the fly},
	url = {https://www.frontiersin.org/articles/10.3389/fnbot.2024.1341750},
	doi = {10.3389/fnbot.2024.1341750},
	language = {English},
	urldate = {2024-05-17},
	journal = {Frontiers in Neurorobotics},
	author = {Engström, Johan and Wei, Ran and McDonald, Anthony D. and Garcia, Alfredo and {Matthew O'Kelly} and Johnson, Leif},
	month = mar,
	year = {2024},
	note = {Publisher: Frontiers},
}

@misc{dinparastdjadid_measuring_2023,
	title = {Measuring {Surprise} in the {Wild}},
	url = {http://arxiv.org/abs/2305.07733},
	doi = {10.48550/arXiv.2305.07733},
	urldate = {2025-01-24},
	publisher = {arXiv},
	author = {Dinparastdjadid, Azadeh and Supeene, Isaac and Engstrom, Johan},
	month = may,
	year = {2023},
	note = {arXiv:2305.07733 [cs]},
}

@inproceedings{fischer_information_2020,
	title = {Information {Particle} {Filter} {Tree}: {An} {Online} {Algorithm} for {POMDPs} with {Belief}-{Based} {Rewards} on {Continuous} {Domains}},
	shorttitle = {Information {Particle} {Filter} {Tree}},
	url = {https://proceedings.mlr.press/v119/fischer20a.html},
	language = {en},
	urldate = {2024-12-16},
	booktitle = {Proceedings of the 37th {International} {Conference} on {Machine} {Learning}},
	publisher = {PMLR},
	author = {Fischer, Johannes and Tas, Omer Sahin},
	month = nov,
	year = {2020},
	note = {ISSN: 2640-3498},
	pages = {3177--3187},
}

@article{laurent_traffic_2021,
	title = {Traffic safety and norms of compliance with rules: {An} exploratory study},
	volume = {41},
	issn = {1545-2921},
	shorttitle = {Traffic safety and norms of compliance with rules},
	url = {http://www.scopus.com/inward/record.url?scp=85125263034&partnerID=8YFLogxK},
	number = {4},
	urldate = {2024-12-11},
	journal = {Economics Bulletin},
	author = {Laurent, Hélène and Sangnier, Marc and Treibich, Carole},
	year = {2021},
	pages = {2464--2483},
}

@article{vasil_world_2020,
	title = {A {World} {Unto} {Itself}: {Human} {Communication} as {Active} {Inference}},
	volume = {11},
	issn = {1664-1078},
	shorttitle = {A {World} {Unto} {Itself}},
	url = {https://www.frontiersin.org/journals/psychology/articles/10.3389/fpsyg.2020.00417/full},
	doi = {10.3389/fpsyg.2020.00417},
	language = {English},
	urldate = {2024-12-02},
	journal = {Frontiers in Psychology},
	author = {Vasil, Jared and Badcock, Paul B. and Constant, Axel and Friston, Karl and Ramstead, Maxwell J. D.},
	month = mar,
	year = {2020},
	note = {Publisher: Frontiers},
}

@article{siebinga_model_2024-1,
	title = {A model of dyadic merging interactions explains human drivers’ behavior from control inputs to decisions},
	volume = {3},
	issn = {2752-6542},
	url = {https://doi.org/10.1093/pnasnexus/pgae420},
	doi = {10.1093/pnasnexus/pgae420},
	number = {10},
	urldate = {2024-11-08},
	journal = {PNAS Nexus},
	author = {Siebinga, Olger and Zgonnikov, Arkady and Abbink, David A},
	month = oct,
	year = {2024},
	pages = {pgae420},
}

@article{montali_waymo_2023,
	title = {The {Waymo} {Open} {Sim} {Agents} {Challenge}},
	volume = {36},
	url = {https://proceedings.neurips.cc/paper_files/paper/2023/hash/b96ce67b2f2d45e4ab315e13a6b5b9c5-Abstract-Datasets_and_Benchmarks.html},
	language = {en},
	urldate = {2024-10-21},
	journal = {Advances in Neural Information Processing Systems},
	author = {Montali, Nico and Lambert, John and Mougin, Paul and Kuefler, Alex and Rhinehart, Nicholas and Li, Michelle and Gulino, Cole and Emrich, Tristan and Yang, Zoey and Whiteson, Shimon and White, Brandyn and Anguelov, Dragomir},
	month = dec,
	year = {2023},
	pages = {59151--59171},
}

@book{parr_active_2022,
	title = {Active {Inference}: {The} {Free} {Energy} {Principle} in {Mind}, {Brain}, and {Behavior}},
	isbn = {978-0-262-04535-3},
	shorttitle = {Active {Inference}},
	language = {en},
	publisher = {MIT Press},
	author = {Parr, Thomas and Pezzulo, Giovanni and Friston, Karl J.},
	month = mar,
	year = {2022},
	note = {Google-Books-ID: UrZNEAAAQBAJ},
}

@article{markkula_defining_2020,
	title = {Defining interactions: a conceptual framework for understanding interactive behaviour in human and automated road traffic},
	shorttitle = {Defining interactions},
	doi = {10.1080/1463922X.2020.1736686},
	journal = {Theoretical Issues in Ergonomics Science},
	author = {Markkula, Gustav and Madigan, Ruth and Nathanael, Dimitris and Portouli, Evangelia and Lee, Yee Mun and Dietrich, Andre and Billington, J and Schieben, Anna and Merat, Natasha},
	year = {2020},
}

@article{markkula_explaining_2023,
	title = {Explaining human interactions on the road by large-scale integration of computational psychological theory},
	volume = {2},
	issn = {2752-6542},
	url = {https://doi.org/10.1093/pnasnexus/pgad163},
	doi = {10.1093/pnasnexus/pgad163},
	number = {6},
	urldate = {2024-04-10},
	journal = {PNAS Nexus},
	author = {Markkula, Gustav and Lin, Yi-Shin and Srinivasan, Aravinda Ramakrishnan and Billington, Jac and Leonetti, Matteo and Kalantari, Amir Hossein and Yang, Yue and Lee, Yee Mun and Madigan, Ruth and Merat, Natasha},
	month = jun,
	year = {2023},
	pages = {pgad163},
}

@article{de_boer_tutorial_2005,
	title = {A {Tutorial} on the {Cross}-{Entropy} {Method}},
	volume = {134},
	issn = {1572-9338},
	url = {https://doi.org/10.1007/s10479-005-5724-z},
	doi = {10.1007/s10479-005-5724-z},
	language = {en},
	number = {1},
	urldate = {2024-04-09},
	journal = {Annals of Operations Research},
	author = {de Boer, Pieter-Tjerk and Kroese, Dirk P. and Mannor, Shie and Rubinstein, Reuven Y.},
	month = feb,
	year = {2005},
	pages = {19--67},
}

@article{friston_active_2017,
	title = {Active {Inference}: {A} {Process} {Theory}},
	volume = {29},
	issn = {0899-7667},
	shorttitle = {Active {Inference}},
	url = {https://doi.org/10.1162/NECO_a_00912},
	doi = {10.1162/NECO_a_00912},
	number = {1},
	urldate = {2023-12-05},
	journal = {Neural Computation},
	author = {Friston, Karl and FitzGerald, Thomas and Rigoli, Francesco and Schwartenbeck, Philipp and Pezzulo, Giovanni},
	month = jan,
	year = {2017},
	pages = {1--49},
}

@article{camara_pedestrian_2021,
	title = {Pedestrian {Models} for {Autonomous} {Driving} {Part} {II}: {High}-{Level} {Models} of {Human} {Behavior}},
	volume = {22},
	issn = {1558-0016},
	shorttitle = {Pedestrian {Models} for {Autonomous} {Driving} {Part} {II}},
	doi = {10.1109/TITS.2020.3006767},
	number = {9},
	journal = {IEEE Transactions on Intelligent Transportation Systems},
	author = {Camara, Fanta and Bellotto, Nicola and Cosar, Serhan and Weber, Florian and Nathanael, Dimitris and Althoff, Matthias and Wu, Jingyuan and Ruenz, Johannes and Dietrich, André and Markkula, Gustav and Schieben, Anna and Tango, Fabio and Merat, Natasha and Fox, Charles},
	month = sep,
	year = {2021},
	pages = {5453--5472},
}

@inproceedings{markkula_evidence_2018,
	address = {Miyazaki, Japan},
	title = {Evidence {Accumulation} {Account} of {Human} {Operators}' {Decisions} in {Intermittent} {Control} {During} {Inverted} {Pendulum} {Balancing}},
	isbn = {978-1-5386-6650-0},
	doi = {10.1109/SMC.2018.00130},
	urldate = {2022-06-15},
	booktitle = {{IEEE} {International} {Conference} on {Systems}, {Man}, and {Cybernetics} ({SMC})},
	author = {Markkula, Gustav and Zgonnikov, Arkady},
	month = oct,
	year = {2018},
	pages = {716--721},
}

@article{schwarting_social_2019,
	title = {Social behavior for autonomous vehicles},
	issn = {0027-8424, 1091-6490},
	doi = {10.1073/pnas.1820676116},
	language = {en},
	urldate = {2019-11-25},
	journal = {Proceedings of the National Academy of Sciences},
	author = {Schwarting, Wilko and Pierson, Alyssa and Alonso-Mora, Javier and Karaman, Sertac and Rus, Daniela},
	month = nov,
	year = {2019},
	pages = {201820676},
}

@article{sadigh_planning_2018,
	title = {Planning for cars that coordinate with people: leveraging effects on human actions for planning and active information gathering over human internal state},
	volume = {42},
	issn = {1573-7527},
	shorttitle = {Planning for cars that coordinate with people},
	doi = {10.1007/s10514-018-9746-1},
	language = {en},
	number = {7},
	urldate = {2019-12-12},
	journal = {Autonomous Robots},
	author = {Sadigh, Dorsa and Landolfi, Nick and Sastry, Shankar S. and Seshia, Sanjit A. and Dragan, Anca D.},
	month = oct,
	year = {2018},
	pages = {1405--1426},
}

@article{pekkanen_variable-drift_2022,
	title = {Variable-{Drift} {Diffusion} {Models} of {Pedestrian} {Road}-{Crossing} {Decisions}},
	volume = {5},
	issn = {2522-087X},
	doi = {10.1007/s42113-021-00116-z},
	language = {en},
	number = {1},
	urldate = {2022-08-24},
	journal = {Computational Brain \& Behavior},
	author = {Pekkanen, Jami and Giles, Oscar Terence and Lee, Yee Mun and Madigan, Ruth and Daimon, Tatsuru and Merat, Natasha and Markkula, Gustav},
	month = mar,
	year = {2022},
	pages = {60--80},
}

@article{lee_road_2021,
	title = {Road users rarely use explicit communication when interacting in today’s traffic: implications for automated vehicles},
	volume = {23},
	issn = {1435-5566},
	shorttitle = {Road users rarely use explicit communication when interacting in today’s traffic},
	doi = {10.1007/s10111-020-00635-y},
	language = {en},
	number = {2},
	urldate = {2022-02-15},
	journal = {Cognition, Technology \& Work},
	author = {Lee, Yee Mun and Madigan, Ruth and Giles, Oscar and Garach-Morcillo, Laura and Markkula, Gustav and Fox, Charles and Camara, Fanta and Rothmueller, Markus and Vendelbo-Larsen, Signe Alexandra and Rasmussen, Pernille Holm and Dietrich, Andre and Nathanael, Dimitris and Portouli, Villy and Schieben, Anna and Merat, Natasha},
	month = may,
	year = {2021},
	pages = {367--380},
}

@article{ljung_aust_fatal_2012,
	title = {Fatal intersection crashes in {Norway}: {Patterns} in contributing factors and data collection challenges},
	volume = {45},
	shorttitle = {Fatal intersection crashes in {Norway}},
	doi = {10.1016/j.aap.2011.11.001},
	journal = {Accident; analysis and prevention},
	author = {Ljung Aust, Mikael and Fagerlind, Helen and Sagberg, Fridulv},
	month = mar,
	year = {2012},
	pages = {782--91},
}

\appendix
\setcounter{equation}{0}
\renewcommand{\theequation}{A.\arabic{equation}}
\newpage
\onecolumn
\section{Appendix}
\subsection{Model implementation}
\label{app:original_model}

In the original model~\cite{schumann_active_2025}, the kinematic state of an agent $v \in V$ is defined as $\bm{x}_v = \{x_v, y_v, \theta_v, \delta_v, v_v\}$ (based on a bicycle model $f$, these are $x$ and $y$ positions, heading angle $\theta$, steering angle $\delta$, and longitudinal velocity $v$) with the kinematic control states $\bm{u}_v = \{a_v, \omega_v\}$ (i.e., longitudinal acceleration $a$ and steering rate $w$). In this work, we expand upon this by adding the signaling states $\bm{\gamma} = \{\gamma_A,\gamma_Y\}$ (prompting and yielding signal respectively). We can then define our relevant variables with
\begin{equation}
    \begin{aligned}
        \bm{\eta}, \bm{o}, \bm{s} & =: \{\bm{x}_v, \bm{\gamma}_v, \bm{u}_v \mid v \in V\} \\
        \bm{a} & =: \{\bm{\gamma}_v, \bm{u}_v \mid v \in V\} \\
        \bm{a}_v & =: \{\bm{\gamma}_v,\bm{u}_v\} \, . \label{eq:state_definition}
    \end{aligned}
\end{equation}
In the following, if it might be unclear if a state comes from, for example, $\bm{o}$ or $\bm{\eta}$, we use the indicators $\bm{x}_{o,v}$ and $\bm{x}_{\eta,v}$.
Based on an implementation of the bicycle model $f_K$~\cite{schumann_active_2025} with 
\begin{equation}
    \bm{x}'_v = f_K(\bm{x}_v, \bm{u}_v, \Delta t) \, ,
\end{equation}
where $\Delta t$ is our discrete simulation timestep, we can then define the \emph{generative process'} probability functions, which as delta distributions $\delta$ can be seen as being deterministic: 
\begin{equation}
    \begin{aligned}
        \widehat{p}(\bm{\eta}' \mid \bm{\eta}, \bm{a}) = & \prod\limits_{v \in V} \delta\left(\bm{x}'_{\eta,v} - f_K(\bm{x}_{\eta,v}, \bm{u}_{a,v}, \Delta t)\right) \delta\left(\bm{\gamma}'_{\eta,v} - \bm{\gamma}_{a,v}\right)  \delta\left(\bm{u}'_{\eta,v} - \bm{u}_{a,v}\right)\\
         \widehat{p}(\bm{o} \mid \bm{\eta}) = &  \prod\limits_{v \in V} \delta\left(\bm{x}_{o,v} - \bm{x}_{\eta,v}\right) \delta\left(\bm{\gamma}_{o,v} - \bm{\gamma}_{\eta,v}\right) \delta\left(\bm{u}_{o,v} - \bm{u}_{\eta,v}\right) \,.
         \label{eq:process_baseline}
    \end{aligned}
\end{equation}
The same can be done for the \emph{generative model} of agent $v$, which will include some more uncertainty, though. Specifically, we have for the state transition function
\begin{equation}
    \begin{aligned}
        p(\bm{s}' \mid \bm{s}, \bm{a}_v) \propto & \, \breve{p}_n(\bm{s}') p_0(\bm{s}' \mid \bm{s}, \bm{a}_v) \\
        p_0(\bm{s}' \mid \bm{s}, \bm{a}_v) = & \, \mathcal{N}\left(\bm{x}'_{s,v} \mid f_K(\bm{x}_{s,v}, \bm{u}_{a,v}, \Delta t), \Sigma_{x,\text{ego}} \right) 
        \delta\left(\bm{\gamma}'_{s,v} - \bm{\gamma}_{a,v} \right) 
        \delta\left(\bm{u}'_{s,v} - \bm{u}_{a,v} \right) \\
        & \prod\limits_{w \in V\setminus{\{v\}}} \mathcal{N}\left(\bm{x}'_{s,w} \mid f_K(\bm{x}_{s,w}, \bm{u}_{s,w}, \Delta t), \Sigma_{x,OV} \right)
        p_\gamma\left(\bm{\gamma}_{s,w}'\vert \bm{\gamma}_{s,w}, \bm{\gamma}_{a,v} \right)
        \mathcal{N}\left(\bm{u}'_{s,w} \mid \bm{u}_{s,w}, \Sigma_{u,OV} \right) \,,
         \label{eq:model_baseline_state}
    \end{aligned}
\end{equation}
where we update the belief about another agents signaling state with the function $p_{\gamma}$, which is defined as
\begin{equation}
        p_\gamma\left(\bm{\gamma}_{s,w}'\vert \bm{\gamma}_{s,w}, \bm{\gamma}_{a,v} \right) = \mathcal{N}_T(\gamma'_{A,s,w} \mid \gamma_{A,s,w}, \sigma_{\gamma}) \begin{cases}
             \mathcal{N}_T(\gamma'_{Y,s,w} \mid  \gamma_{Y,s,w}, \sigma_{\gamma}) & \gamma_{A,a,v} = 0 \\
             \mathcal{B}_{\gamma_{Y,s,w}}(\gamma'_{Y,s,w}) & \gamma_{A,a,v} = 1
        \end{cases}     \,.  \label{eq:signal_state_transition}
\end{equation}
Here, $\mathcal{B}$ is the Bernoulli distribution, while $\mathcal{N}_T$ is a truncated normal distribution with bounds in $(0,1)$. It must be noted that theoretically using a Bernoulli distribution to map into a continuous space is problematic, but as it would be possible to approximate the discrete Bernoulli distribution for instance with a weighted average of two truncated normal distributions (Equation~\eqref{eq:bernoulli_approx}) without any major effect on model behavior, this implementation is not a major issue.
\begin{equation} \label{eq:bernoulli_approx}
    \underset{\epsilon \rightarrow 0}{\lim} \left( \gamma_{Y,s,w} \mathcal{N}_T(\gamma'_{Y,s,w} \mid  1, \varepsilon) + (1-\gamma_{Y,s,w}) \mathcal{N}_T(\gamma'_{Y,s,w} \mid  0, \varepsilon) \right) = \mathcal{B}_{\gamma_{Y,s,w}}(\gamma'_{Y,s,w}) 
\end{equation}

Meanwhile, a part of this specific active inference model is the \emph{projected normative probability} $\breve{p}_n$~\cite{schumann_active_2025}, which is used to bias an agent towards the belief that other road users will behave according to established traffic rules~\cite{schumann_active_2025,laurent_traffic_2021, abbas_drivers_2024} encoded in the \emph{normative probability} $p_n$. It is defined as
\begin{equation}
    \breve{p}_n(\bm{s}') = \min \left\{p_n(\bm{s}'), \left( \frac{1}{\vert\breve{T}\vert} \sum\limits_{t \in \breve{T}} p_n(\breve{\bm{s}}'^{(t)})^{-1} \right)^{-1} \right\} \, . \label{eq:state_norm}
\end{equation}
where $\breve{\bm{s}}'^{(i)}$ is generated by using $p_0$ (Equation~\eqref{eq:model_baseline_state}) to propagate forward $\bm{s}$ for $i$ timesteps, using a given policy. Importantly, however, during this rollout, we assume that $\Sigma_{u,OV} = \bm{0}$, to minimize noise accumulation. While in the baseline of the model one assumes $\breve{T} = \{1, H_n\}$, we here use $\breve{T} = \{1, \frac{1}{2}(1 + H_n), H_n\}$ instead.

Meanwhile, the observation function is more simplistic.
\begin{equation}
    \begin{aligned}
         \widehat{p}(\bm{o} \mid \bm{\gamma}) & = \delta\left(\bm{o} - \bm{\eta}\right)\\
         p(\bm{o} \mid \bm{s}) & =\mathcal{N}\left(\bm{x}_{o,v} \mid \bm{x}_{s,v}, \Sigma_{x,o}\right) \,
         \mathcal{N}\left(\bm{u}_{o,v} \mid \bm{u}_{s,v}, \Sigma_{u,o} \right)\,
         \mathcal{B}_{\bm{\gamma}_{s,v}}\left(\bm{\gamma}_{o,v}\right) \\
         & \hphantom{=} \prod\limits_{w \in V\setminus{\{v\}}} \mathcal{N}\left(\bm{x}_{o,w} \mid \bm{x}_{s,w}, \Sigma_{x,o}\right) \,
         \mathcal{N}\left(\bm{u}_{o,w} \mid \bm{u}_{s,w}, \Sigma_{u,o} \right)\,
         \mathcal{B}_{\bm{\gamma}_{s,w}^{10}}\left(\bm{\gamma}_{o,w}\right) 
         \label{eq:model_baseline_obs}
    \end{aligned}
\end{equation}
Here, the use of the power $10$ in the Bernoulli distribution ensures that positive signals result in a much greater change to the current belief than negative ones, to represent the idea that a given signal (representing a deeper, lasting intent) will not be immediately forgotten.

\subsubsection{Minimizing Variational Free Energy} \label{sec:Belief_update}
Fundamentally to the active inference framework, each agent tries minimize its \emph{free energy}, which first effect the updating of the agent's belief $q(\bm{s})$ (minimizing \emph{variational free energy}). The latter is based on the existences of a preference prior $p(\bm{o})$ depicting an agents preferred observations.
To minimize its \emph{variational free energy} (VFE) an agent first makes an observation $\bm{o}_t$ of the world state $\bm{\eta}_t$ at each timestep $t$, with
\begin{equation}
    \bm{o}_t \sim \widehat{p}(\bm{o}'\vert \bm{\eta}_{t})\,.
\end{equation}
With this observation, the agent can update its old belief $q(\bm{s}_{t-1})$ using Bayesian inference~\cite{friston_active_2017, engstrom_resolving_2024}, with 
\begin{equation}\begin{aligned} \label{eq:variation_inference_orig}
    q(\bm{s}_{t}) & \propto p(\bm{o}_{t}\vert \bm{s}_{t}) \, \underbrace{\mathbb{E}_{q(\bm{s}_{t-1})} p(\bm{s}_{t}\vert \bm{s}_{t-1}, \bm{a}_{v, t-1})}_{q_A(\bm{s}_{t})}\,.
\end{aligned}\end{equation}

Following the collision avoidance model~\cite{schumann_active_2025}, we use a particle filter approach to implement the belief $q(\bm{s})$, which is consequently represented by a set of $N$ samples $\bm{S} = \{\bm{s}_1, \hdots, \bm{s}_N\}$. Consequently, to determine $q(\bm{s}_t)$, we need to generate the corresponding set $\bm{S}_t$. In a first step, we advance individual particles $\bm{S}_{t-1}$ using the transition function to the expected states $\bm{S}_{A, t} = \{\bm{s}_{A, t, n} \sim p(\bm{s}'\vert \bm{s}_{t - 1, n}, \bm{a}_{v, t-1}) \mid n \in \{1,\hdots, N\}\} $ representing $q_A(\bm{s}_{t}) $. Based on the expected state $\bm{S}_{A, t}$, we can represent this expected belief distribution $q_A(\bm{s}_{t})$ with a kernel density estimate~\cite{fischer_information_2020}, which we want to be the conjugate prior for the observation likelihood $p(\bm{o}_{t}\vert \bm{s}_{t})$:
\begin{itemize}
    \item For non-communication states $\bm{x}$ and $\bm{u}$, as the corresponding dimensions in the observation probability are Gaussian (Equation~\eqref{eq:model_baseline_obs}), we use Gaussian kernels.
    \item For communication states $\bm{\gamma}$, where the corresponding observations use a Bernoulli likelihood (Equation~\eqref{eq:model_baseline_obs}), we use beta-distribution kernels. For the cases with the power $10$, a positive update is therefore treated as 8 consecutive positive observations, while a negative update is treated as $0.15$ consecutive negative observations ($\mathcal{B}_{\gamma_s^{10}}(\gamma=0) = 1 - \gamma_s^{10}$ is approximated with $(1- \gamma_s)^{M_{\mathcal{B}}}$ the closest for $M_{\mathcal{B}} = 0.15$). Therefore, a positive update has greater impact on the respective belief $q(\gamma_s)$ than a negative one.
\end{itemize}
Consequently, using conjugate priors, we can then generate a kernel density distribution for representing the posterior belief $q(\bm{s}_{t})$ consisting of similar types of kernels (with different parameters and weights) will then take the form of a Gaussian mixture model. Sampling from those will be elementary.

\subsubsection{Minimizing Expected Free Energy}
\label{app:EFE}
The minimization of free energy is also reflected in the policy selection, where an agent $v$ selects actions $\bm{a}_{v,t}$ that minimize the \emph{expected free energy} (EFE)~\cite{friston_active_2017,engstrom_resolving_2024}. These actions are determined as part of finding a longer policy $\bm{\pi}_{v,t} = \{\bm{a}_{v,\tau} \mid \tau \in \{t, \hdots, t + H - 1\}\} $ with prediction horizon $H$ that minimizes the EFE $G$, defined as
\begin{equation}
    G(\bm{\pi}_{v,t}) = \sum\limits_{\tau = t + 1}^{t + H}  - g_{\text{prag}}\left(\widetilde{q}(\bm{o}_{\tau})\right) - g_{\text{epist}}\left(\widetilde{q}(\bm{s}_{\tau}) \right) \, ,
    \label{eq:EFE_definition}
\end{equation}
and combining a \emph{pragmatic value} $g_{\text{prag}}$ and an \emph{epistemic value} $g_{\text{epist}}$, with 
\begin{equation}\begin{aligned}
    g_{\text{prag}}\left(\widetilde{q}(\bm{o})\right) & = \mathbb{E}_{\widetilde{q}(\bm{o})} \ln p(\bm{o}) \\
    g_{\text{epist}}\left(\widetilde{q}(\bm{s}) \right) & = \underbrace{\mathcal{H}(\mathbb{E}_{\widetilde{q}(\bm{s})} p(\bm{o}\vert \bm{s}))}_{\text{Posterior predictive entropy}}  - \underbrace{\mathbb{E}_{\widetilde{q}(\bm{s})} \mathcal{H}(p(\bm{o}\vert \bm{s}))}_{\text{Expected Ambiguity}} \, , \label{eq:pragmatic_epistemic}
\end{aligned} 
\end{equation}
where $\mathcal{H}$ is the Shannon entropy~\cite{friston_active_2017}. For the epistemic value $g_{\text{epist}}$, the first term represents the uncertainty about the the future observations resulting from the current policy, while the second one represent the ambiguity associated with those future observations~\cite{engstrom_resolving_2024}. In maximizing the epistemic value, the agent therefore tries to generate policies that minimize that ambiguity in observations as varied as possible.
In those equations, $\widetilde{q}(\bm{s}_{\tau}) = \mathbb{E}_{\widetilde{q}(\bm{s}_{\tau - 1})} p(\bm{s}_{\tau}\vert \bm{s}_{\tau - 1}, \bm{a}_{v,\tau - 1})$ is the predicted state belief and $ \widetilde{q}(\bm{o}_{\tau}) = \mathbb{E}_{\widetilde{q}(\bm{s}_{\tau})} p(\bm{o}_{\tau}\vert \bm{s}_{\tau})$ the predicted observation belief. These are expressed by the sets of particles $\widetilde{\bm{S}}_{\tau}$  and $\widetilde{\bm{O}}_{\tau}$ (representing $\widetilde{q}(\bm{s}_{\tau})$ and $\widetilde{q}(\bm{o}_{\tau})$, respectively), which are generated with 
\begin{equation}
\begin{aligned}
    \widetilde{\bm{S}}_{\tau} & = \{ \bm{s}_{\tau,n} \sim p(\bm{s}' \vert \bm{s}_{\tau - 1, n}, \bm{a}_{v,\tau - 1}) \mid n \in \{1, \hdots, N \}\} \\
    \widetilde{\bm{O}}_{\tau} & = \{\bm{o}_{\tau,n} \sim p(\bm{o}'\vert \bm{s}_{\tau,n}) \mid n \in \{1,\hdots, N\} \}  \, .
\end{aligned}
\end{equation}
Based on this particle filter, the parts of the EFE (Equation~\eqref{eq:pragmatic_epistemic}) can be calculated as 
\begin{equation} \label{eq:pragmatric_implement}
    g_{\text{prag}}\left(\widetilde{\bm{O}}\right) = \frac{1}{\vert \widetilde{\bm{O}}\vert}\sum\limits_{\bm{o} \in \widetilde{\bm{O}}} \ln p(\bm{o})
\end{equation}
and
\begin{equation}\begin{aligned}
    g_{\text{epist}}\left(\widetilde{\bm{S}},\widetilde{\bm{O}}\right) = & \, - \frac{1}{N} \sum\limits_{\bm{o} \in \widetilde{\bm{O}}} \ln \left( \frac{1}{N} \sum\limits_{\bm{s} \in \widetilde{\bm{S}}}  p(\bm{o}\vert \bm{s}) \right) - \frac{1}{N} \sum\limits_{\bm{s} \in \widetilde{\bm{S}}} \mathcal{H}\left(p(\bm{o}'\vert \bm{s}) \right) \, , \label{eq:epistemic_real}
\end{aligned}
\end{equation}
assuming that the entropy $\mathcal{H}\left(p(\bm{o}'\vert \bm{s}) \right) = H_p(\bm{s})$ is analytically calculable. Meanwhile, the specific implementation of the preference prior $p(\bm{o})$ is discussed below.
The EFE $G$ is then minimized by the \emph{cross entropy method}~\cite{de_boer_tutorial_2005}, whose use in an active inference model has been previously described in detail~\cite{engstrom_resolving_2024, schumann_active_2025}. 

\subsubsection{Accumulating surprise}
A crucial finding of Schumann \emph{et al.}~\cite{schumann_active_2025} was the need for delaying the generation of a new policy $\bm{\pi}_{v,t}$ to allow the active inference model to produce human-like reaction times. 

Tot this end, the agent is continually accumulating \emph{surprise} $\epsilon$ (here defined as defined as the \emph{residual information} of the pragmatic value~\cite{dinparastdjadid_measuring_2023}), with  
\begin{equation}\label{eq:accumulated_surprise}
    E_t =E_{t-1} +  \lambda\,\, \underbrace{\left( H \underset{\bm{o}}{\max} \ln p(\bm{o}) - \sum\limits_{\tau = t + 1}^{t+H} \mathbb{E}_{\widetilde{q}(\bm{o}_{\tau})} \ln p(\bm{o}_\tau) \right)}_{\epsilon_t}\, .
\end{equation}
If the accumulated evidence $E_t$ does not surpass the threshold of $1$, the agent just extends the unexecuted parts of its previous policy $\bm{\pi}_{v,t-1}$ with a further action, while a new policy is created otherwise
\begin{equation}
    \bm{\pi}_{v,t}^* = \begin{cases}
        \left(\bm{\pi}_{v,t-1}\setminus \{\bm{a}_{v,t-1}\}\right) \cup \underset{\{\bm{a}_{v,t+H-1}\}}{\text{argmin}} \, G(\{\bm{a}_{v,t+H-1}\}) & E_t < 1 \\
        \underset{\bm{\pi}_{v,t}}{\text{argmin}} \, G(\bm{\pi}_{v,t}) & E_t \geq 1
    \end{cases}   \, .
\end{equation}
The latter option of generating a new policy is also referred to as re-planning~\cite{schumann_active_2025}.

\subsubsection{Preference priors}\label{sec:pref_prior}
In the previous model~\cite{schumann_active_2025}, we define the preference prior $p(\bm{o})$ (underlying the \emph{pragmatic value}, see Equation~\eqref{eq:pragmatic_epistemic}) for a modeled vehicle $v$ as
\begin{equation}\label{eq:preference_likelihood}
    \begin{aligned}
        p(\bm{o}) = & \; \mathcal{N}(v_{o,v}\vert \mu_v, \sigma_v) \, \mathcal{N}(a_{o,v}\vert 0, \sigma_a) \mathcal{N}(\omega_{o,v}\vert 0, \sigma_\omega) \, p_{\text{lat}}(y_{o,v}) \, p_{\text{coll}}(\bm{o})\, p_{\text{safe}}(\bm{o}) \,.
    \end{aligned}
\end{equation}
Those components and their corresponding parameters are mostly kept intact. However, we generalize the lane keeping aspect to $p_{\text{lat}}(d_{\text{lat},o,v})$ with $d_{\text{lat},o,v} = y_{o,v} \cos(\theta_{\text{road}}) - x_{o,v} \sin (\theta_{\text{road}})$ to allow for cases were the lane boundaries are not aligned with $\theta_{\text{road}}=0$ (a further generalization for cases where the lane centerline does not cross the origin could be considered, but was not required in our modeled scenario from Figure~\ref{fig:overview}).

Nevertheless, we also added some additional components, that will be multiplied to the preference prior:
\begin{itemize}
    \item \textbf{Speed limit:} The speed limit preference prior discourages agents from driving too fast. It is implemented as
    \begin{equation}
        p_{\text{SL}}(v_{o,v}) =\begin{cases}
             \exp\left(g_S \frac{v_{o,v} - \SI{10}{m.s^{-1}}}{\SI{4.2}{m.s^{-1}}} \right)& v_{o,v} > \SI{10.278}{m.s^{-1}} \\
            1 & \text{Otherwise}
        \end{cases}  
    \end{equation}
    \item \textbf{Stop sign:} For implementing the stop sign rules, for each agent, we implement an additional boolean $h_v$, which indicates if the vehicle has stopped already or not. In the generative model's state transition function, it is updated for the ego agent $v$ with
    \begin{equation}
        p(h'_{s,v}\mid \bm{s}, \bm{a}_v) = \delta\left(h'_{s,v} -f_h(\bm{s}' \sim p(\bm{s}' \mid \bm{s},\bm{a}_v), h_{s,v}, v)\right)
    \end{equation}
    with the state transition function defined in Equation~\eqref{eq:model_baseline_state} and 
    \begin{equation}
    \begin{aligned}
        f_h(\bm{s}, h_{s,v}, v) = h_{s,v} \lor \left(\left(\SI{-19.425}{m} \leq d_{\text{long},s,v} \leq \SI{-3.925}{m}\right) \land \left(v_v < \SI{0.278}{m.s^{-1}}\right)\right)
    \end{aligned}
    \end{equation}
    where
    \begin{equation}
        d_{\text{long},s,v}' = x_{s,v}' \cos(\theta_{\text{road}}) + y_{s,v}' \sin (\theta_{\text{road}}) \,.
    \end{equation}
    Additionally, in the observation function, we assume that $p(h_{o,v}|h_{s,v}) = \delta(h_{o,v}-h_{s,v})$.
    We can then define the corresponding preference term
    \begin{equation}
    p_{\text{stop}}(\bm{o}) = \begin{cases}
        \exp(g_S) &  (\SI{-3.925}{m} \leq d_{\text{long},o,v}) \land \lnot h_{o,v} \\
        1 & \text{Otherwise}
    \end{cases}
    \end{equation}
    \item \textbf{Priority rule:} The preference prior implementing the ``first come, first go'' priority rule is interactive, i.e., similar to the collision loss, it depends on both the state of the ego agent $v$ and the other agent $w$. For that, we add an additional boolean state $p_v$ which indicates if agent $v$ assumes that it has priority. Its state transition function is defined as 
    \begin{equation}
        p(p'_{s,v}\mid \bm{s}, \bm{a}_v) = \delta\left(p'_{s,v} - f_{\text{lead}}(\bm{s}' \sim p(\bm{s}' \mid \bm{s},\bm{a}_v), \bm{s}, p_{s,v}, v, w)\right) \,.
    \end{equation}
    Here, the function $f_{\text{lead}}(\bm{s}',\bm{s},p_{s,v},v,w)$ assigns priority to agent $v$ over agent $w$ (if so, return 1, otherwise, return 0), depending on the existence of stop signs. If those exist, then we define 
    \begin{equation}
        f_{\text{lead}}(\bm{s}',\bm{s},p_{s,v},v,w) = p_{s,v} \lor \left(h'_{s,v} \land \lnot h_{s,v} \land \left(d_{\text{long},s,v}' > d_{\text{long},s,w}'\right) \right) \,.
    \end{equation}
    Without stop signs, this becomes
    \begin{equation}
        f_{\text{lead}}(\bm{s}',\bm{s},p_{s,v},v,w) = p_{s,v} \lor \left(\left(d_{\text{long},s,v} < \SI{-20}{m} \leq d_{\text{long},s,v}'\right)  \land\left(\frac{d_{\text{long},s,v}'}{\max\{0,v_{s,v}\}} > \frac{d_{\text{long},s,w}'}{\max\{0,v_{s,w}\}}\right) \right) \,.
    \end{equation}
    Similar to the stop rule, in the observation function, we assume that $p(p_{o,v}|p_{s,v}) = \delta(p_{o,v}-p_{s,v})$.
    We can then define the corresponding preference prior:
    \begin{equation}
        p_{\text{priority}}(\bm{o}) = \begin{cases}
            \exp\left(\frac{1}{2}g_S \right) & \left(d_{\text{long},o,v} > \max\{\SI{-3.925}{m},d_{\text{long},o,w} - \SI{4.425}{m}\} \right) \land \lnot p_{o,v} \\
            1 & \text{Otherwise}
        \end{cases}
    \end{equation}
    Assuming that communication is enabled, this function is slightly adjusted so that yielding by the other agent can override priority rules:
    \begin{equation}
        p_{\text{priority}}(\bm{o}) = \begin{cases}
            \exp\left(\frac{1}{2}g_S \right) & \left(d_{\text{long},o,v} > \max\{\SI{-3.925}{m},d_{\text{long},o,w} - \SI{4.425}{m}\} \right) \land \lnot p_{o,v} \land \lnot \gamma_{Y,o,w}\\
            1 & \text{Otherwise}
        \end{cases}
    \end{equation}
    \item \textbf{Signaling:} For signaling there are two different aspects to consider. The first one is the general cost of signaling
    \begin{equation}
        p_{comm}(\bm{o}) = \exp\left(g_{\gamma} \left(\gamma_{A,o,v} + \gamma_{Y,o,v} \right)\right) \,.
    \end{equation}
    If there are stop signs, it is instead adjusted to prevent signaling before stopping:
    \begin{equation}
        p_{comm}(\bm{o}) = \exp\left(g_{\gamma} \left(\gamma_{A,o,v} + \gamma_{Y,o,v} \right) \begin{cases}
            1 & h_{o,v} \\
            10 & \text{Otherwise}
        \end{cases}\right) \,.
    \end{equation}
    Meanwhile, there is also a preference prior against uncooperative agents, punish both signaling of yielding without the intention to yield and the lack of a yield signal if intending to yield and prompting by the other agent $w$.
    This would be expressed as
    \begin{equation}
        p_{\text{coop}}(\bm{o}) = \begin{cases}
            \exp\left(g_{W}\right) & \left(\gamma_{Y,o,v} \land p_{o,v}\right) \lor \left( \gamma_{A,o,w} \land \lnot p_{o,v} \land \lnot \gamma_{Y,o,v} \right)  \\
            1 & \text{Otherwise}
        \end{cases}
    \end{equation}
    In cases without priority rules, using the sigmoid function $\sigma$, we assume that
    \begin{equation}
        p_{o,v} \sim \mathcal{B}_{\sigma\left(3\left(\frac{d_{\text{long},s,v}}{\max\{0,v_{s,v}\}} - \frac{d_{\text{long},s,w}}{\max\{0,v_{s,w}\}}\right)\right)}
    \end{equation}
\end{itemize}

\subsubsection{Normative likelihood}\label{sec:norm_prob}
In the normative likelihood $p_n(\bm{s})$ (used in the state transition function, see Equation~\eqref{eq:state_norm}), we adjust the functions from the collision avoidance model~\cite{schumann_active_2025} enforcing lane following with the following terms (multiplied together):
\begin{itemize}
    \item \textbf{Speed limit:} In regard to speeding of the other vehicle $w$, we assume that $p_n$ is multiplied with 
    \begin{equation}
        p_{n,\text{SL}}(\bm{s}) = \begin{cases}
            0.02 & v_{s,w} > \SI{11.5}{m.s^{-1}} \\
            1 & \text{Otherwise}
        \end{cases} \, .
    \end{equation}
    \item \textbf{Stop sign:}
    If stop signs exist, we use the further multiplier
    \begin{equation}
        p_{n,\text{stop}}(\bm{s}) = \begin{cases}
            0.02 & (\SI{-3.925}{m} \leq d_{\text{long},o,w}) \land \lnot h_{s,w}  \\
            1 & \text{Otherwise}
        \end{cases} \, .
    \end{equation}
    \item \textbf{Priority rule:} Meanwhile, if the priority rule is enabled, the ego agent $v$ assigns the other agent $w$ the following normative probability multiplier:
    \begin{equation}
        p_{n,\text{priority}}(\bm{s}) = \begin{cases}
            0.02 & \left(d_{\text{long},s,w} > \max\{\SI{-3.925}{m},d_{\text{long},s,v} - \SI{4.425}{m}\} \right) \land p_{s,v}  \\
            1 & \text{Otherwise}
        \end{cases} \, .
    \end{equation}
    \item \textbf{Signaling:}
    If communication is enabled, the model is biased to believe that the other agent will follow up on their signaled yielding, as long as this belief is certain enough ($\gamma_{Y,s,w}>0.2$):
    \begin{equation}
        p_{n,\text{coop}}(\bm{s}) = \begin{cases}
            1- 1.4\max\{\gamma_{Y,s,w}-0.3, 0\} & d_{\text{long},s,w} > \max\{\SI{-3.925}{m},d_{\text{long},s,v} - \SI{4.425}{m}\}  \\
            1 & \text{Otherwise}
        \end{cases} \, .
    \end{equation}
\end{itemize}

\subsection{Model parameters}
While we mostly reuse the parameters from Table~1 from the original model~\cite{schumann_active_2025}, some variables are changed, and new ones are added, which can be seen in Table~\ref{tab:Parameters}. The change that stands out most here is the different value implementing the speed preference $\sigma_v$, which was decreased to prevent the potential of deadlocks in scenarios where norms or communication are sufficient to remove the risk of a collision, but where the cross entropy method approach is unable to converge to the superior approach of accelerating across the intersection. This change, however, should not be seen as major, as the parameter sensitivity analysis in the previous model~\cite{schumann_active_2025} showed no significant difference in performance under such a change.

\begin{table*}
    \centering
    \caption{An overview of the new and changed parameters compared to the original model (see Table~1 in Schumann \emph{et al.}~\cite{schumann_active_2025}).}
    \begin{tabular}{|p{0.2\textwidth}|p{0.425\textwidth}|p{0.325\textwidth}|}
    \hline
    \rowcolor{black!66}
    \textcolor{white}{\textbf{Parameter}} & \textcolor{white}{\textbf{Description}} & \textcolor{white}{\textbf{Source}} \\ \hline
    \rowcolor{black!33}
    \multicolumn{3}{|l|}{State transition function $p(\bm{s}' \vert \bm{s}, \bm{a})$} \\ \hline
    $\sigma_{\gamma,0} = 0.005$ & Noise applied to other vehicle’s signalling during belief update & Tuned parameter \\ \hline
    $\sigma_{\gamma} = 0.001$ & Noise applied to other vehicle’s signalling during behavior prediction & Tuned parameter \\ \hline
    \rowcolor{black!33}
    \multicolumn{3}{|l|}{Preference function $p(\bm{o})$} \\ \hline
    $\sigma_v = 0.2 $ [\unit{m.s^{-1}}] & Standard deviation of preference distribution of agent's velocity $v$ & Tuned parameter \\ \hline
    $\sigma_a = 0.2$ [\unit{m.s^{-2}}] & Standard deviation of preference distribution of agent's acceleration $a$ & Tuned parameter \\ \hline
    $g_{S} = - 10000$ & Pragmatic value of norm violation & Tuned parameter \\ \hline
    $g_{\gamma} = - 0.125$ & Pragmatic value of an active communication signal & Tuned parameter  \\ \hline
    $g_{W} = - 10000$ & Pragmatic value of uncooperative signalling & Tuned parameter  \\ \hline
    \rowcolor{black!33}
    \multicolumn{3}{|l|}{Policy selection} \\ \hline
    $\mathcal{N}\left(0, 10^{-5} \right)$ [\unit{s^{-1}}] & Distribution to sample steering rates $\omega_{\tau}$ in first CEM iteration -- set very low to avoid steering responses & Previous model~\cite{schumann_active_2025} (mean) and Tuned parameter (std)  \\ \hline
    $\lambda = 10 ^ {-5.9} $ & Evidence accumulation drift rate & Tuned parameter \\ \hline
         
    \end{tabular}
    \label{tab:Parameters}
\end{table*}

\subsection{The epistemic value of prompting}
In our model, an agent $v$ chooses to prompt the other agent $w$ for their \emph{yielding intent} ($\gamma_{A,a,v} =1$), if the epistemic value advantage of that action is bigger than the pragmatic cost $g_{\gamma}$ associated with signaling. We can show that this overall advantage (here defined as $g_{\text{prompt}}$) mostly depends on the current predictive posterior entropy of this action.
\begin{equation}\begin{aligned}\label{eq:g_prompt}
    g_{\text{prompt}}(q(\gamma_{Y,s,w})) & =  g_{\text{epist}}(\widetilde{q}(\gamma_{Y,s,w}\vert\gamma_{A,a,v} =1)) - g_{\text{epist}}(\widetilde{q}(\gamma_{Y,s,w}\vert\gamma_{A,a,v} =0)) - g_{\gamma}
\end{aligned}
\end{equation}
Here, assuming that $q(\gamma_{Y,s,w}) \approx \delta (\gamma_{Y,s,w} - p_{\text{yield}})$,
we can find the following using the state transition function (Equation~\eqref{eq:signal_state_transition}):
\begin{equation}
    \begin{aligned}
        \widetilde{q}(\gamma_{Y,s,w}'\vert\gamma_{A,a,v} =1)& = \mathbb{E}_{q(\gamma_{Y,s,w})} p(\gamma_{Y,s,w}'\vert \gamma_{Y,s,w}, \gamma_{A,a,v}) \\
        & = \mathbb{E}_{\delta (\gamma_{Y,s,w} - p_{\text{yield}})} \mathcal{B}_{\gamma_{Y,s,w}}(\gamma_{Y,s,w}') \\
        & = \mathcal{B}_{p_{\text{yield}}}(\gamma_{Y,s,w}') \\
        \widetilde{q}(\gamma_{Y,s,w}'\vert\gamma_{A,a,v} =0)& = \mathbb{E}_{q(\gamma_{Y,s,w})} p(\gamma_{Y,s,w}'\vert \gamma_{Y,s,w}, \gamma_{A,a,v}) \\
        & = \mathbb{E}_{\delta (\gamma_{Y,s,w} - p_{\text{yield}})} \mathcal{N}_T(\gamma_{Y,s,w}'\vert \gamma_{Y,s,w}, \sigma_y) \\
        & \approx \delta (\gamma_{Y,s,w}' - p_{\text{yield}})\\
    \end{aligned}
\end{equation}
Here, we ignore the influence of the noise $\sigma_y$, as for one timestep, it is negligible.
Using the definition of the epistemic value in Equation~\eqref{eq:pragmatic_epistemic} (where, using Equation~\eqref{eq:model_baseline_obs} we assume that $p(\bm{o}\vert\bm{s}) = p(\gamma_{Y,o,w}\vert \gamma_{Y,s,w}) = \mathcal{B}_{\gamma_{Y,s,w} ^ {10}}(\gamma_{Y,o,w})$), we can find that Equation~\eqref{eq:g_prompt} simplifies to:
\begin{equation}\begin{aligned} \label{eq:g_prompt_full}
    g_{\text{prompt}}(q(\gamma_{Y,s,w})) & =  g_{\text{epist}}(\widetilde{q}(\gamma_{Y,s,w}\vert\gamma_{A,a,v} =1)) - g_{\text{epist}}(\widetilde{q}(\gamma_{Y,s,w}\vert\gamma_{A,a,v} =0)) - g_{\gamma}  \\
    & \approx  g_{\text{epist}}(\mathcal{B}_{p_{\text{yield}}}(\gamma_{Y,s,w})) - g_{\text{epist}}(\delta (\gamma_{Y,s,w} - p_{\text{yield}})) - g_{\gamma}  \\
    & \approx \left( \underbrace{\mathcal{H}\left(\mathbb{E}_{\mathcal{B}_{p_{\text{yield}}}(\gamma_{Y,s,w})} \mathcal{B}_{\gamma_{Y,s,w} ^ {10}}(\gamma_{Y,o,w})\right)}_{\text{PPE}}  - \mathbb{E}_{\mathcal{B}_{p_{\text{yield}}}(\gamma_{Y,s,w})} \underbrace{\mathcal{H}\left(\mathcal{B}_{\gamma_{Y,s,w} ^ {10}}(\gamma_{Y,o,w})\right)}_{\mathcal{H}(\mathcal{B}_0(\gamma))=\mathcal{H}(\mathcal{B}_1(\gamma)) = 0}\right)\\ & \hphantom{=}\, - \left( \mathcal{H}\left(\mathbb{E}_{\delta (\gamma_{Y,s,w}' - p_{\text{yield}})} \mathcal{B}_{\gamma_{Y,s,w} ^ {10}}(\gamma_{Y,o,w})\right)  - \mathbb{E}_{\delta (\gamma_{Y,s,w}' - p_{\text{yield}})} \mathcal{H}\left(\mathcal{B}_{\gamma_{Y,s,w} ^ {10}}(\gamma_{Y,o,w})\right)\right) - g_{\gamma} \\
    & \approx \underbrace{\mathcal{H}\left(\mathcal{B}_{p_{\text{yield}}}(\gamma_{Y,o,w})\right)}_{\text{PPE}}  - \underbrace{\left( \mathcal{H}\left( \mathcal{B}_{p_{\text{yield}}^ {10}}(\gamma_{Y,o,w})\right)  - \mathcal{H}\left(\mathcal{B}_{p_{\text{yield}}^ {10}}(\gamma_{Y,o,w})\right)\right)}_{g_{\text{epist}}(\widetilde{q}(\gamma_{Y,s,w}\vert\gamma_{A,a,v} =0)) \approx 0} - g_{\gamma}\\
    & \approx \underbrace{\mathcal{H}\left(\mathcal{B}_{p_{\text{yield}}}(\gamma_{Y,o,w})\right)}_{\text{PPE}} - g_{\gamma}
\end{aligned}
\end{equation}

Consequently, it can be easily seen that $g_{\text{prompt}}(q(\gamma_{Y,s,w})\approx \delta (\gamma_{Y,s,w} - p_{\text{yield}}))$ is maximized for $p_{\text{yield}} \approx 0.5$.
It should also be noted that in our implementation, the pragmatic cost for prompting might be greater in certain circumstances (for example, if an agent is already signaling yielding).

\end{document}